\documentclass{tlp}
\usepackage{amsmath}
\usepackage{latexsym}
\usepackage{epsfig}
\usepackage{psfrag}
\usepackage{verbatim}

\newcommand{\algop}[2]{\underset{#2}{#1}}

\def\vect#1{{\bf #1}}

\def\email#1{{\tt#1}}

\newcommand{\nop}[1]{}

\newcommand{\myfigure}[3]{
	\begin{figure}[tbp]
	\includegraphics[width=0.8\textwidth]{#3}
	\caption{\label{#1}#2}
	\end{figure}
}

\def\relation#1{{\it #1}}
\def\dom{\relation{DOM}}
\def\fail{\relation{FAIL}}
\def\found{\relation{FOUND}}
\def\guess{{\it Guess}}
\def\npalg{{\it NP-Alg}}

\def\ampl{{\sc ampl}}
\def\opl{{\sc opl}}

\def\sql{{\sc sql}}
\def\consql{{\sc conSql}}
\def\consqlsim{{\sc conSql simulator}}
\def\jlocal{{\sc JLocal}}

\newtheorem{definition}{Definition}[section]
\newtheorem{theorem}{Theorem}[section]
\newtheorem{lemma}{Lemma}[section]

\begin{document}

  \title[Combining Relational Algebra, \sql{}, Constraint Modelling, and Local Search]
        {Combining Relational Algebra, \sql{}, Constraint Modelling, and Local Search\protect{\footnote{This paper is an extended and revised version of \cite{cado-manc-02}.}}
	}
  \author[Marco Cadoli and Toni Mancini]
         {MARCO CADOLI and TONI MANCINI\\
         Dipartimento di Informatica e Sistemistica\\
	 Universit\`a di Roma ``La Sapienza''\\
	 Via Salaria 113, 00198 Roma, ITALY\\
         \email{cadoli|tmancini@dis.uniroma1.it}}

\submitted{15 December 2003}
\revised{15 April  2005}
\accepted{5 January 2006}

\maketitle

\begin{abstract}
  The goal of this paper is to provide a strong integration between constraint modelling and relational DBMSs. To this end we propose extensions of standard query languages such as relational algebra and \sql{}, by adding constraint modelling capabilities to them. 
  In particular, we propose non-deterministic extensions of both languages, which are specially suited for combinatorial problems.
  Non-determinism is introduced by means of a \emph{guessing} operator, which
  declares a set of relations to have an arbitrary extension.  This new
  operator results in languages with higher expressive power, able to express
  all problems in the complexity class NP.  Some syntactical restrictions which
  make data complexity polynomial are shown.  The effectiveness of both
  extensions is demonstrated by means of several examples.
  The current implementation, written in Java using local search techniques, is described.
\end{abstract}


\section{Introduction}
\label{sec:intro}
The efficient solution of NP-hard combinatorial problems, such as resource
allocation, scheduling, planning, etc. is crucial for many industrial
applications, and it is often achieved by means of ad-hoc procedural hand-written
programs. Declarative programming languages like \ampl\ \cite{four-gay-kern-93} and \opl\ \cite{vanh-99} or libraries \cite{ILOG98} for expressing
constraints are commercially available.
Data encoding the instance are either in text files in an ad-hoc
format, or in standard relational DBs accessed through libraries callable from
programming languages such as C++  (cf., e.g., \cite{ILOG-DBLINK}).
In other words, there is not a strong integration between data definition
and constraint modelling and programming languages.

Indeed, such an integration is particularly needed in industrial environments, where the necessity for solving combinatorial problems coexists with the presence of large databases where data to be processed lie. 
Hence, constraint solvers that operate externally to the databases of the enterprise may lead to a series of disadvantages, first of all a potential lack of the integrity of recorded data.
To this end, a better coupling between standard data repositories and constraint solving engines is highly desiderable.

The goal of this paper is exactly to integrate constraint modelling and programming into relational
database management systems (R-DBMSs).
In particular, we show how standard query languages for relational databases can be extended in order to give them constraint modelling and solving capabilities: with such languages, constraint problem specifications can be viewed just like (more complex) queries to standard data repositories. 
In what follows, we propose extensions of standard query languages such as relational algebra and \sql, 
that are able to formulate queries defining combinatorial and constraint problems.

%%%%%
In principle relational algebra can be used as a language for testing constraints. As an
example, given relations $A$ and $B$, testing whether all tuples in $A$ are
contained in $B$ can be done by computing the relation $A-B$, and then checking
its emptiness.  Anyway, it must be noted that relational algebra is unfeasible as a language
for expressing NP-hard problems, since it is capable of expressing just a
strict subset of the polynomial-time queries (cf., e.g.,
\cite{abit-hull-vian-95}).  As a consequence, an extension is needed.
%%%%%

The proposed generalization of relational algebra is named \npalg, and it is proven to be
capable of expressing all problems in the complexity class NP. We focus on NP
because this class contains the decisional version of most combinatorial
problems of industrial relevance \cite{garey-johnson:79:book}.  \npalg\ is relational algebra plus a
simple \emph{guessing} operator, which declares a set of relations to have an
arbitrary extension.  Algebraic expressions are used to express constraints.
Several interesting properties of \npalg\ are provided: its data complexity is
shown to be NP-complete, and for each problem $\xi$ in NP we prove that there
is a fixed query that, when evaluated on a database representing the instance
of $\xi$, solves it. Combined complexity is also addressed.

Since \npalg\ expresses all problems in NP, an interesting question is whether
a query corresponds to an NP-complete or to a polynomial-time problem. We give a
partial answer to it, by exhibiting some syntactical restrictions of \npalg\
with polynomial-time data complexity.

In the same way, \consql\ (\sql\ with constraints) is the proposed non-deterministic extension of \sql,
the well-known language for querying relational databases
\cite{ullman:88:principles-vol1}, having the same expressive power of \npalg, and supporting also the specification of optimization problems.
We believe that writing a \consql\ query for the solution of a combinatorial optimization
problem is only moderately more difficult than writing \sql{} queries for a
standard database application. The advantage of using \consql\ is twofold: it is
not necessary to learn a completely new language or methodology, and
integration of the problem solver with the information system of the enterprise
can be done very smoothly.  The effectiveness of both \npalg\ and \consql\ as
constraint modelling languages is demonstrated by showing several queries which
specify combinatorial and optimization problems.

The structure of the paper is as follows.
Syntax and semantics of \npalg\ are introduced in Section~\ref{sec:np-alg}.
Some examples of \npalg\ queries for the specification of NP-complete
combinatorial problems are proposed in Section~\ref{sec:examples}.
Main computational properties of \npalg, including data and combined
complexity, expressive power, and polynomial fragments, are presented in
Section~\ref{sec:computational}. Section~\ref{sec:consql} contains some
details of \consql\ and its implementation \consqlsim, as well as the specification of some
real-world combinatorial and optimization problems.
Finally, Section~\ref{sec:related} contains conclusions as
well as references to main related work.

\section{\npalg: Syntax and semantics}
\label{sec:np-alg}
We refer to a standard definition of relational algebra with the five operators $\{\sigma, \pi,
\times, -, \cup\}$ \cite{abit-hull-vian-95}. Other operators such as
``$\Join$'' and ``$\slash$'' can be defined as usual.  
Attributes (fields) of relations will be denoted either by their names or by their indexes. As an example, given a relation $\relation{R}(a,b)$, the selection of tuples in \relation{R} with the same values for the two attributes will be denoted in one of the following forms: $\algop{\sigma}{\relation{R.a}=\relation{R.b}}(\relation{R})$, $\algop{\sigma}{\relation{a}=\relation{b}}(\relation{R})$ (since there is no confusion to what relation \relation{a} and \relation{b} refer to), $\algop{\sigma}{\$1=\$2}(\relation{R})$.
As for join conditions, they will have atoms of the form $\relation{a}=\relation{b}$ or $\relation{a}\neq \relation{b}$ where $\relation{a}$ is an attribute name (or even index) of the relation on the left of the join symbol, and $\relation{b}$ one of that on the right.
Finally, temporary relations such
as $T = algexpr(\ldots)$ will be used to make expressions easier to read. As
usual
(cf., e.g., \cite{chan-hare-80})
queries are defined as mappings which
are partial recursive and generic, i.e., constants are uninterpreted.

Let $D$ denote a finite relational database, $\vect{R}$ the set of its relations,
and \dom\ the unary relation representing the set of all constants occurring in
$D$.

\begin{definition}[Syntax of \npalg]
\label{def:npalg-syn}
An \npalg\ expression has two parts:
\begin{enumerate}
\item A set $\vect{Q} = \{Q_1^{(a_1)},\ldots,Q_n^{(a_n)}\}$ of new relations of
  arbitrary arity, denoted as $\guess\ Q_1^{(a_1)},\ldots,Q_n^{(a_n)}$.
 Sets $\vect{R}$ and $\vect{Q}$ must be disjoint.
\item An ordinary expression $exp$ of relational algebra on the new database schema
  $[\vect{Q},\vect{R}]$.
\end{enumerate}
\end{definition}

\noindent
For simplicity, until Section~\ref{sec:computational} we focus on \emph{boolean queries}, i.e., queries that admit a yes/no answer.
For this reason we restrict $exp$ to be a relation which we call \fail.

\begin{definition}[Semantics of \npalg]
\label{def:npalg-sem}
The semantics of an \npalg\ expression is as follows:
\begin{enumerate}
\item For each possible extension $ext$ of the relations in $\vect{Q}$ with elements in
 \dom, the relation \fail\ is evaluated, using ordinary rules of relational algebra.
\item If there exists an extension $ext$ such that the expression for \fail\ evaluates to the empty relation ``$\emptyset$'' (denoted as $\fail\Diamond\emptyset$), the
  \emph{answer} to the boolean query is ``yes''. Otherwise the \emph{answer} is ``no''.
  
  When the answer is ``yes'', the extension of relations in $\vect{Q}$ is a
  \emph{solution} for the problem instance.
\end{enumerate}
\end{definition}

\noindent
A trivial implementation of the above semantics obviously requires exponential
time, since there are exponentially many possible extensions of the relations in
$\vect{Q}$. Anyway, as we will show in Section~\ref{sec:polynomial}, some
polynomial-time cases indeed exist.

The reason why we focus on a relation named \fail\ is that, typically, it is
easy to specify a decision problem as a set of constraints (cf.\ forthcoming
Sections~\ref{sec:examples} and \ref{sec:consql}). As a consequence, an
instance of the problem has a solution if and only if there is an arbitrary choice of the
guessed relations such that all constraints are satisfied, i.e.,
$\fail=\emptyset$.  A $\found^{(1)}$ query can be anyway defined as
$
\found = \dom - \algop{\pi}{\$1}(\dom \times \fail)$.
In this case, the answer is ``yes'' if and only if there is an extension $ext$ such
that $\found\neq\emptyset$.

\section{Examples of \npalg\ queries}
\label{sec:examples}

In this section we show the specifications of some NP-complete problems, as
queries in \npalg.  All examples are on uninterpreted structures, i.e., on
unlabeled directed graphs, because we adopt a pure relational algebra with uninterpreted
constants.  As a side-effect, the examples show that, even in this limited
setting, we are able to emulate bounded integers and ordering.  This is very important,
because the specification of very simple combinatorial problems requires
bounded integers and ordering.

In Section~\ref{sec:consql} we use the full power of \consql\ to
specify some real-world problems.

\subsection{Graph $k$-coloring}
\label{sec:coloring}
We assume a directed graph is represented as a pair of relations $\relation{NODES}^{(1)}(n)$
and $\relation{EDGES}^{(2)}(from, to)$ (with tuples in $\relation{EDGES}^{(2)}$ having components in $\relation{NODES}^{(1)}$, hence, $\dom = \relation{NODES}$).  A graph is
\emph{$k$-colorable} if there is a \emph{k}-partition
$Q_1^{(1)},\ldots,Q_k^{(1)}$ of its nodes, i.e., a set of $k$ sets such
that:
\begin{itemize}
\item $\forall i \in [1,k], \forall j \in [1,k], j \neq i \rightarrow Q_i \cap Q_j = \emptyset$,

\item $\bigcup_{i=1}^k Q_i = \relation{NODES}$,
\end{itemize}
and each set $Q_i$  has no pair of nodes linked by an edge.
The problem is well-known to be NP-complete for $k \geq 3$ (cf., e.g., \cite{garey-johnson:79:book}),
and it can be specified in \npalg\ as follows:

\begin{subequations}
\begin{align}
&\guess\ Q_1^{(1)},\ldots, Q_k^{(1)}; \label{kCol-Guess}\\
&\fail\relation{\_DISJOINT} = \bigcup_{ 
        \begin{subarray}{c}
		i\neq j \in \{1,\ldots,k\}
        \end{subarray} 
        }
        Q_i \algop{\Join}{  
        } Q_j; \label{kCol-Disg}\\
&\fail\relation{\_COVER} =\relation{NODES} \ \Delta \ \bigcup_{i=1}^k Q_i;
\label{kCol-Copr}\\
&\fail\relation{\_PARTITION} = 
\label{kCol-Part} \fail\relation{\_DISJOINT} \ \cup \ \fail\relation{\_COVER}; \\
&\fail\relation{\_COLORING} = \label{kCol-Coloraz}
  \algop{\pi}{\$1} \left[
        \bigcup_{i=1}^k \left(
        \left( \algop{\sigma}{\$1\neq\$2}(Q_i \times Q_i) \right)
                \algop{\Join}{
                        \begin{subarray}{c}
                                \$1=\relation{EDGES}.from\\
                                \$2=\relation{EDGES}.to
                        \end{subarray} 
                } \relation{EDGES}
        \right)
        \right];
        \\[2mm]
&\fail =   \label{kCol-Fail}  \fail\relation{\_PARTITION} \ \cup \ \fail\relation{\_COLORING}.
\end{align}
\end{subequations}

\noindent
Expression \eqref{kCol-Guess} declares \emph{k} new relations of arity 1.
Expression \eqref{kCol-Fail} collects all constraints a candidate coloring must
obey to:
\begin{itemize}
\item \eqref{kCol-Disg} and \eqref{kCol-Copr} make sure that $Q_1,\ldots,Q_k$ is a partition
  of $\relation{NODES}$ (``$\Delta$'' is the symmetric difference operator, i.e., $A \ 
  \Delta \ B = \left( A-B \right) \cup \left( B-A \right)$, useful for testing
  equality since $A \ \Delta \ B = \emptyset \Longleftrightarrow A = B$).
  
\item \eqref{kCol-Coloraz} checks that each set $Q_i$ has no pair of nodes
  linked by an edge.
\end{itemize}

\noindent
As an example, let $k= 3$ and the database be as follows: \nopagebreak[4]

\bigskip

\begin{small}
\begin{tabular}{cc}
\begin{minipage}[t]{2cm}
\begin{center}
\noindent
\relation{NODES} \medskip

\noindent
		\begin{tabular}{|c|}
		\cline{1-1} 
 		\relation{n} \\
 		\cline{1-1}
 		1 \\ 2 \\ 3 \\ 4 \\
 		\cline{1-1}
 		\end{tabular}
\end{center}
\end{minipage}
&
\begin{minipage}[t]{3cm}
\begin{center}
\noindent
\relation{EDGES} \medskip
	
\noindent
		\begin{tabular}{|c|c|}
		\cline{1-2}
		\relation{from} & \relation{to} \\
		\cline{1-2}
		1 & 2\\
		1 & 4 \\
		2 & 3 \\
		\cline{1-2}
		\end{tabular}
\end{center}
\end{minipage}
\end{tabular}
\end{small}

\bigskip

\noindent
An extension of $\relation{Q}_1$, $\relation{Q}_2$ and $\relation{Q}_3$ such that $\fail = \emptyset$ is:

\nopagebreak[4]
\bigskip

\begin{small}
\begin{tabular}{ccc}
\begin{minipage}[t]{2cm}
\begin{center}

$\relation{Q}_1$ \medskip

		\begin{tabular}{|c|}
		\cline{1-1}
 		--- \\
 		\cline{1-1}
		2 \\ 4 \\
 		\cline{1-1}
 		\end{tabular}
\end{center}
\end{minipage}
&
\begin{minipage}[t]{2cm}
\begin{center}
$\relation{Q}_2$ \medskip

		\begin{tabular}{|c|}
		\cline{1-1}
 		--- \\
 		\cline{1-1}
		1 \\
 		\cline{1-1}
 		\end{tabular}
\end{center}
\end{minipage}
&
\begin{minipage}[t]{2cm}
\begin{center}
$\relation{Q}_3$ \medskip

		\begin{tabular}{|c|}
		\cline{1-1}
 		--- \\
 		\cline{1-1}
		3 \\
 		\cline{1-1}
 		\end{tabular}
\end{center}
\end{minipage}
\end{tabular}
\end{small}

\bigskip

\noindent
Note that such an extension constitutes a solution to the coloring problem.

We observe that in the specification above the
\fail\relation{\_PARTITION} relation \eqref{kCol-Part} makes sure that an extension of
 $Q_1^{(1)},\ldots,$ $Q_k^{(1)}$ is a \emph{k-partition} of \relation{NODES}.
Such a constraint 
can be very useful for the specification of problems, so we
introduce, as syntactic sugar, an expression:
\begin{equation*}
\relation{failPartition}^{(1)}(N^{(k)}, P_1^{(k)}, \ldots, P_n^{(k)}),
\end{equation*}
which returns an empty relation if and only if $\{P_1^{(k)}, \ldots, P_n^{(k)}\}$ is a
partition of $N^{(k)}$. The prefix \relation{fail} in the name of the 
expression
reminds the user that it should be used in checking constraints. 
We note that the arity of \relation{failPartition} can, without loss of generality, be fixed to 1, since we can always project out the remaining columns.
Other 
useful syntactic sugar
will be introduced in the following examples, and 
is
summarized in
Section~\ref{sec:syntsugar}.

\subsection{Independent set}
\label{sec:indset}
Let a (directed) graph be defined, as usual, by the two relations
$\relation{NODES}^{(1)}$ and $\relation{EDGES}^{(2)}$, and let $k \leq |\relation{NODES}|$ be an integer, which
is specified by a relation $\relation{K}^{(1)}$ containing exactly $k$ tuples. A
subset \relation{N} of \relation{NODES}, with $|\relation{N}| \geq k$ is said to be an \emph{independent set of size at least $k$} of the graph if \relation{N} contains no pair of nodes linked by an edge.

The problem of determining whether an input graph has an independent set of
size at least $k$ is NP-complete (cf., e.g., \cite{garey-johnson:79:book}), and it can
be easily specified in \npalg. 
However, since we have to compare the size of \relation{N} with the integer $k$ (i.e., with the size of relation \relation{K}), before presenting the \npalg\ query for the Independent set problem, we
need a method to compare the size of two relations $\relation{N}^{(1)}$ and $\relation{K}^{(1)}$. This can be done by deciding whether a proper \emph{function} that maps tuples in \relation{N} to tuples in \relation{K} exists. In particular:

\begin{itemize}
\item $|\relation{N}| = |\relation{K}|$ if and only if there exists a total bijective function between \relation{N} and \relation{K};

\item $|\relation{N}| \geq |\relation{K}|$ if and only if there exists a partial surjective function from \relation{N} to \relation{K};

\item $|\relation{N}| \leq |\relation{K}|$ if and only if there exists a total injective function from \relation{N} to \relation{K}.
\end{itemize}

\noindent
To define relational algebra expressions that check whether a relation $\relation{FUN}^{(d+r)}$ is a (total, injective, surjective, or bijective) \emph{function} from domain $\relation{D}^{(d)}$ to range $\relation{R}^{(r)}$, we define the following 
expressions
(for the sake of simplicity, we write definitions for $d=r=1$, but their extensions to arbitrary $d$ and $r$ are straightforward):
\begin{itemize}
\item $\relation{failFunction}(\relation{FUN}^{(2)},\relation{D}^{(1)},\relation{R}^{(1)}) =$
\[
           \left( \algop{\pi}{\$1}(\relation{FUN})\ - \ \relation{D} \right)
                                \ \cup \
                                \left( \algop{\pi}{\$2}(\relation{FUN})\ - \ \relation{R} \right) \ \cup \
		\algop{\pi}{\$1} \left(
			\relation{FUN} 
				\algop{\Join}{
					\begin{subarray}{c}
						\$1 = \$1 \\*
                                                \land \\*
                                                \$2 \neq \$2
					\end{subarray}
                                }
			\relation{FUN}
                                         \right),
\]
where the first and second subexpressions check whether tuples in \relation{FUN} are in the cartesian product $\relation{D}\times\relation{R}$, and the third checks whether \relation{FUN} is mono-valued;

\item $\relation{failTotal}(\relation{FUN}^{(2)},\relation{D}^{(1)},\relation{R}^{(1)}) \ = \
           \relation{D} \ - \ \algop{\pi}{\$1}\left(\relation{FUN}\right)$;

\item $\relation{failSurjective}(\relation{FUN}^{(2)},\relation{D}^{(1)},\relation{R}^{(1)}) \ = \
           \relation{R} \ - \ \algop{\pi}{\$2}\left(\relation{FUN} \right)$;

\item $\relation{failInjective}(\relation{FUN}^{(2)},\relation{D}^{(1)},\relation{R}^{(1)}) \ = \
	\relation{FUN} 
		\algop{\Join}{
			\begin{subarray}{c}
				\$1 \neq \$1 \\*
				\land \\*
                                \$2 = \$2
			\end{subarray}
                 } 
		 \relation{FUN}	$.
	   
\end{itemize}

\noindent
The above 
expressions
evaluate to the empty relation if and only if relation \relation{FUN} is, respectively, a function, a total, surjective, or injective relation (in the mathematical sense) from tuples of relation \relation{D} to tuples of relation \relation{R}.

By using the above 
expressions,
we can design new ones, with the goal of comparing the size of two relations \relation{D} and \relation{R}:
\begin{itemize}
\item $\relation{failGeqSize}(\relation{AUX},\relation{D},\relation{R}) = \relation{failFunction}(\relation{AUX},\relation{D},\relation{R}) \cup
\relation{failSurjective}(\relation{AUX},\relation{D},\relation{R})$;

\item $\relation{failLeqSize}(\relation{AUX},\relation{D},\relation{R}) = 
\relation{failGeqSize}(\relation{AUX},\relation{R},\relation{D})$;

\item $\relation{failEqSize}(\relation{AUX},\relation{D},\relation{R}) = 
\relation{failLeqSize}(\relation{AUX},\relation{D},\relation{R}) \cup
\relation{failGeqSize}(\relation{AUX},\relation{D},\relation{R})$,
\end{itemize}
where \relation{AUX} is an auxiliary guessed relation that encodes the function between \relation{D} and \relation{R}.
Such auxiliary guessed relations will be omitted as arguments in the remainder of the paper if they are not used anywhere else, to enhance readability.

\bigskip \noindent
Returning to the example, the following \npalg\ query 
specifies the Independent set problem: 

\begin{align*}
&\guess\ \relation{N}^{(1)}; \\*
&\fail = \begin{aligned}[t]
                &\relation{failGeqSize}^{(1)}(\relation{N},\relation{K}) \ \cup
		\\*
                %&
                &\algop{\pi}{\$1} \Big[
                        (\relation{N} \times \relation{N}) \algop{\Join}{
                                \begin{subarray}{c}
                                        \$1 = \relation{EDGES}.from \\
                                        \land \\
                                        \$2 = \relation{EDGES}.to
                                \end{subarray}
                                }
                                \relation{EDGES}
                        \Big].
              \end{aligned}
\end{align*}

\noindent
The first subexpression of \fail\ specifies the constraint $|\relation{N}| \geq k$, while the second one evaluates to the empty relation if and only if no pair of nodes in \relation{N} is linked by an edge. An extension of \relation{N} is an
independent set (with size at least $k$) of the input graph if and only if the
corresponding \fail\ relation is empty.

\subsection{Clique}
\label{sec:clique}
Given an undirected graph, i.e., the \relation{EDGES} relation is symmetric,
 and an integer $k \leq |\relation{NODES}|$, a subset
\relation{N} of \relation{NODES}, with $|\relation{N}| \geq k$ is said to be a \emph{clique of size at least  $k$} if every pair of distinct nodes of \relation{N} is linked by an edge
(i.e., the subgraph induced by \relation{N} is \emph{complete}).

The problem of determining whether a graph has a clique of size at least $k$ is
NP-complete (cf., e.g., \cite{garey-johnson:79:book}), and it can be specified in
\npalg\ as follows ($k$ is encoded as a relation $\relation{K}^{(1)}$ with exactly $k$
tuples):

\begin{align*}
&\guess\ \relation{N}^{(1)}; \\*
&\fail = %%\begin{aligned}[t]
                %&
               \relation{failGeqSize}(\relation{N},\relation{K}) %\ 
               \cup %%%\\*
                %&
               \Big( \algop{\sigma}{\$1 \neq \$2} (\relation{N} \times \relation{N})
                        \algop{\Join}{  \begin{subarray}{c}
                                                \$1 = \$1 \\
                                                \land \\
                                                \$2 = \$2
                                        \end{subarray}
                        }
                        \relation{complement}^{(2)}(\relation{EDGES})
              %\end{aligned}
             \Big).
\end{align*}

\noindent
The structure of the query is very similar to the one of the previous example,
except for the new 
expression
$\relation{complement}^{(k)}(\relation{R}^{(k)})$, which can be defined as
\begin{equation*}
complement^{(k)}(\relation{R}^{(k)}) = \algop{\rho}{\begin{subarray}{c}
                                                \$1 \to \relation{R}.\$1 \\
                                                \vdots \\
                                                \$k \to \relation{R}.\$k
                                         \end{subarray}
                            } (\dom^k - \relation{R}),
\end{equation*}
and returns the \emph{active complement} of the relation given as argument
($\rho$ is the field-renaming operator, used to name all columns of the output relation like those of $\relation{R}$). 
Obviously the above query can be written in several other ways. As an example, a more efficient one would use the difference operator, instead of the join;
notwithstanding this, we have chosen the above query to show the use of \relation{complement}.

\subsection{More examples}
\label{sec:moreexamples}
We can specify in \npalg\ other famous problems over graphs like
\emph{Dominating set}, \emph{Transitive closure}, and \emph{Hamiltonian
 path}.
It is worth noting that Transitive closure, indeed a polynomial-time problem, is
not expressible in relational algebra (cf., e.g.,~\cite{abit-hull-vian-95}), because it
intrinsically requires a form of recursion (cf.\ Section~\ref{sec:related}).
In \npalg\ recursion can be simulated by means of guessing. 
As for Hamiltonian path, this is the problem of finding a traversal of a graph which touches each node
exactly once.
The possibility to specify the Hamiltonian path problem in \npalg\ has interesting consequences which deserve some comments.  
Consider a unary relation \dom, with $|\dom| = M \neq 0$ and the complete graph	\relation{C}
defined by
the relations $\relation{NODES} = \dom$ and $\relation{EDGES} = \dom \times \dom$.
A Hamiltonian path \relation{H} of \relation{C} is a total ordering of the $M$ elements in \dom: in
fact it is a \emph{successor} relation.
The transitive closure of \relation{H} is the corresponding \emph{less-than} relation.
As a consequence, we have the possibility to use \emph{bounded integers} in the range $[1,M]$ in our framework, and also arithmetic operations on them.

Furthermore, the Hamiltonian paths of \relation{C} correspond to the \emph{permutations} of $[1,M]$.
Permutations are very useful for the specification of several problems.
As an example, in the $n$-\emph{queens} problem (in which the goal is to place $n$
non-attacking queens on an $n\times n$ chessboard) a candidate solution is a
permutation of order $n$, representing the assignment of a pair $\langle \text{row},
\text{column}\rangle$ to each queen. Interestingly, to check the attacks of queens
on diagonals, in \npalg\ we can guess a relation encoding the subtraction of
elements in \dom.

Other interesting problems, not involving graphs, can be specified in \npalg: \emph{Satisfiability of a propositional formula} and \emph{Evenness of the cardinality of a relation} are some examples.

\subsection{Useful syntactic sugar}
\label{sec:syntsugar}
Previous examples show that guessing relations as subsets of $\dom^k$ (for integer $k$) is enough to express many NP-complete problems.
The forthcoming Theorem~\ref{thm:expressiveness} shows that this is indeed enough
to express all problems in NP.

Nevertheless, 
expressions
such as \relation{failPartition} can make queries more
readable.  In this section we briefly summarize the  main 
expressions
that we
designed.

\begin{itemize}
  
\item $\relation{empty}^{(1)}(\relation{R}^{(k)}) = \dom - \algop{\pi}{\$1} ( \dom \times \relation{R}^{(k)} )$, evaluates to the empty relation if \relation{R} is a non-empty one (and vice versa).
  
\item $\relation{complement}^{(k)}(\relation{R}^{(k)})$ 
  evaluates to the active complement (with respect to $\dom^k$) of \relation{R} (cf.\ Section~\ref{sec:clique}).
  
\item $\relation{failPartition}^{(1)}(\relation{N}^{(k)}, \relation{P}_1^{(k)}, \ldots, \relation{P}_n^{(k)})$ (cf.\ 
  Section~\ref{sec:coloring}) evaluates to the empty relation if and only if
  $\{\relation{P}_1^{(k)},\ldots,\relation{P}_n^{(k)}\}$ is a partition of \relation{N}.

\item $\relation{failSuccessor}^{(1)}(\relation{SUCC}^{(2k)}, \relation{N}^{(k)})$ evaluates to the empty relation if and only if \relation{SUCC} encodes a correct successor relation on elements in \relation{N}, i.e., a 1-1 correspondence with the interval $[1,|\relation{N}|]$ (essentially by checking whether \relation{SUCC} is a Hamiltonian path on the graph with edges defined by $\relation{N} \times \relation{N}$).

\item $\relation{failPermutation}^{(1)}(\relation{PERM}^{(2k)},N^{(k)})$ evaluates to the empty relation if and only if \relation{PERM} is a permutation of the elements in $\relation{N}^{(k)}$.  The ordering sequence is given by the first $k$ columns of \relation{PERM}.

\item \sloppy{
$\relation{failFunction}^{(1)}(\relation{FUN}^{(d+r)}, \relation{D}^{(d)}, \relation{R}^{(r)})$, 
$\relation{failTotal}^{(1)}(\relation{FUN}^{(d+r)}, \relation{D}^{(d)}, \relation{R}^{(r)})$, 
$\relation{failInjective}^{(1)}(\relation{FUN}^{(d+r)}, \relation{D}^{(d)}, \relation{R}^{(r)})$,  
$\relation{failSurjective}^{(1)}(\relation{FUN}^{(d+r)}, \relation{D}^{(d)}, \relation{R}^{(r)})$  
}
(cf. Section~\ref{sec:indset})
evaluate to the empty relation if and only if \relation{FUN} is, respectively, a function, a total, injective or surjective relation from tuples in \relation{D} to those in \relation{R}.
We remark that, since elements in \relation{R} can be ordered (cf.\ Section~\ref{sec:moreexamples}), \relation{FUN} is also an \emph{integer function} from elements of \relation{D} to the interval
  $[1,|\relation{R}|]$. 
  Integer functions are very useful for the specification of \emph{resource
   allocation} problems, such as \emph{Integer knapsack} (see also examples in
  Section~\ref{sec:sqlexamples}).

\item $\relation{failEqSize}^{(1)}(\relation{N},\relation{K})$, $\relation{failGeqSize}^{(1)}(\relation{N},\relation{K})$,
  $\relation{failLeqSize}^{(1)}(\relation{N},\relation{K})$ (cf.\ Section~\ref{sec:indset}) evaluate to the empty relation if and only if $|\relation{N}|$ is, respectively, $=$, $\geq$, $\leq$ $|\relation{K}|$.
\end{itemize}

\section{Computational aspects of \npalg}
\label{sec:computational}
In this section we focus on the main computational aspects of \npalg: data and
combined complexity, expressive power, and polynomial fragments.

Technically, the results presented in this section can be easily obtained from corresponding ones formulated for other languages, e.g., existential second order logic (ESO). Nevertheless, we believe that when designing a constraint modelling language it is of fundamental importance, from the methodological point of view, to ascertain its main computational properties.

\subsection{Data and combined complexity}
\label{sec:complexity}
The \emph{data complexity}, i.e., the complexity of query answering assuming
the data\-ba\-se as input and a fixed query (cf.\ \cite{abit-hull-vian-95}),
is one of the
most important computational aspects of a language, since queries are
typically small compared to the database.

Since we can express some NP-complete problems in \npalg\ (cf.\
Section~\ref{sec:examples}), the problem of deciding whether
$\fail\Diamond\emptyset$ is NP-hard.
Moreover we can prove that the data complexity for such a problem is in NP by
using the following argument.  It is possible to generate, in non-deterministic
polynomial time, an extension $ext$ of $Q$.  The answer is ``yes'' if and only
if there is such an $ext$ that makes $\fail = \emptyset$. The last check, being
the evaluation of an ordinary relational algebra expression, can be done in polynomial time in
the size of the database.
The above considerations give us the first computational result on \npalg.

\begin{theorem}
\label{thm:datacompl}
The data complexity of deciding whether $\fail\Diamond\emptyset$ for an
\npalg\ query, where the input is the database, is NP-complete.
\end{theorem}

\noindent
Another interesting measure is \emph{combined complexity}, where both the
database and the query are part of the input.  
It is well known that, typically, the combined complexity of a language is much higher than its data complexity~\cite{vardi:82:complexity}.
As for \npalg, it is possible to show that, when both the database and the query are part of the input, the problem of determining whether $\fail\Diamond\emptyset$ is hard for the
complexity class NE, defined as $\bigcup_{c>1}  NTIME~(2^{cn})$ (cf.\
\cite{papadimitriou:94:computational}), i.e., the class of all problems solvable by a
non-determin\-ist\-ic machine in time bounded by $2^{cn}$, where $n$ is the size of
the input and $c$ is an arbitrary constant.

\begin{theorem}
\label{thm:combinedcompl}
The combined complexity of deciding whether $\fail\Diamond\emptyset$ for an
\npalg\ query, where the input is both the database and the query, is NE-hard.
\end{theorem}

\noindent
The proof is quite long and is delayed to \ref{sec:proofCombined}.

\subsection{Expressive power}
\label{sec:expressiveness}
The \emph{expressiveness} of a query language characterizes the problems
that can be expressed as fixed, i.e., instance independent, queries.  In this
section we prove the main result about the expressiveness of \npalg, by showing
that it captures exactly NP, or equivalently (cf.\ \cite{fagi-74}) queries in
the existential fragment of second-order logic (ESO).

Of course it is very important to be assured that we can express \emph{all}
problems in the complexity class NP. In fact, Theorem~\ref{thm:datacompl} says
that we are able to express \emph{some} problems in NP.  We remind that the
expressive power of a language is less than or equal to its data
complexity. In other words, there exist languages whose data complexity is hard for
class $C$ in which not every query in $C$ can be expressed;  several such
languages are known, cf., e.g., \cite{abit-hull-vian-95}.

In order to show that \npalg\ is able to express all problems in NP, we 
illustrate a method that transforms an arbitrary formula in ESO into a \npalg\ query. 
We remind that, by Fagin's theorem
\cite{fagi-74}, 
any collection $\vect{D}$ of finite databases over
$\vect{R}$ is NP-recognizable if and only if it can be defined by a existential second order formula.
In particular, we deal with ESO formulae of the following kind:

\begin{equation}
%\label{eq:npalg-consql:eso_ae}
\label{SO}
(\exists \vect{S}) \ (\forall \vect{X}) \ (\exists \vect{Y})  \ \varphi(\vect{X},\vect{Y}),
\end{equation}

\noindent
where $\varphi$ is a first-order formula (without quantifiers) containing variables among $\vect{X},\vect{Y}$ and involving relational symbols in $\vect{S} \cup \vect{R} \cup \{ =
\}$.  
The reason why we can restrict our attention to second-order formulae in
the above normal form is explained in \cite{kola-papa-91}.  As usual, ``='' is
always interpreted as ``identity''.

The transformation works in two steps:
\begin{enumerate}
\item \label{point:translationOfPhi}
The first-order formula $\varphi(\vect{X},\vect{Y})$ obtained by
  eliminating all quantifiers from~\eqref{SO} is translated into 
  an expression
  \relation{PHI} of plain relational algebra;

\item The query $\psi$ is defined as:
\begin{equation} \label{eqn:theorem}
  \begin{split}
	&\guess\ Q_1^{(a_1)},\ldots,Q_n^{(a_n)}; \\
	&\fail = \dom^{|\vect{X}|} - \algop{\pi}{\vect{X}}(\relation{PHI}),
  \end{split}
\end{equation}
\end{enumerate}
where $a_1,\dots,a_n$ are the arities of the $n$ predicates in $\vect{S}$, and
$|\vect{X}|$ is the number of variables occurring in $\vect{X}$.

The first step is rather standard (cf., e.g., \cite{abit-hull-vian-95}), and is briefly sketched here just to give the intuition. 
A relation \relation{R} (with the same arity) is introduced for each
predicate symbol $r$ in the relational vocabulary of $\varphi$, i.e., $\vect{R} \cup \vect{S}$.  An atomic formula of first-order
logic is translated as the corresponding relation, possibly prefixed by a
selection that accounts for constant symbols and/or repeated variables,
and by a renaming of attributes mapping the arguments.
Selection can be used also for dealing with atoms involving equality.
Inductively, the relation corresponding to a complex first-order formula is
built as follows:

\begin{itemize}
\item $f \land g$ translates into $F \algop{\Join}{} G$, where $F$ and $G$
  are the translations of $f$ and $g$, respectively;
\item $f \lor g$ translates into $F' \cup G'$, where $F'$ and $G'$ are
  derived from the translations $F$ and $G$ to account for the (possibly)
  different schemata of $f$ and $g$;
\item $\lnot f(\vect{Z})$ translates into $\algop{\rho}{\begin{subarray}{c}
                                                \$1 \to F.\$1 \\
                                                \vdots \\
                                                \$|\vect{Z}| \to F.\$|\vect{Z}|
                                         \end{subarray}
                            } (\dom^{|\vect{Z}|} - F)$
($\rho$ is the column renaming operator, needed to name columns of $(\dom^{|\vect{Z}|} - F)$ like those of $F$).
\end{itemize}

\noindent
It is worth noting that a better translation avoids the insertion of occurrences of the \dom\ relation for the important class of \emph{safe formulae} (cf., e.g., \cite{abit-hull-vian-95}). However, these issues are out of the scope of this paper, and will not be taken into account.

Relations obtained through such a translation will be called $q$\emph{-free}, 
because they do not contain the projection operator (that plays the role of an existential \emph{quantification}), except those implicit in
equi-joins. Intuitively, this means that there are no existential quantifiers.

The following theorem claims that the above translation is correct.
\begin{theorem}
\label{thm:expressiveness}
For any NP-recognizable collection $\vect{D}$ of finite databases over $\vect{R}$
 --characterized by a formula of the kind \eqref{SO}--
a database $D$ is in $\vect{D}$, i.e., $D\models (\exists \vect{S})(\forall
 \vect{X})(\exists\vect{Y}) ~   \varphi(\vect{X},\vect{Y})$,
  if and only if $\fail\Diamond\emptyset$, when $\psi$ (cf.\ formula
  \eqref{eqn:theorem}) is evaluated on $D$.

\begin{proof}
(Only if part) 
If $D \in \vect{D}$, it follows that an extension $\vect{\Sigma}$ for predicates in $\vect{S}$ exists, such that:
\begin{equation*}
	[D, \vect{\Sigma}] \models (\forall \vect{X}) (\exists \vect{Y}) ~ \varphi (\vect{X}, \vect{Y}). 
\end{equation*}

\noindent
By translating $\varphi$ on the right side into relational algebra (according to the point~\ref{point:translationOfPhi} above), we obtain a relational expression \relation{PHI} on the relational vocabulary given by relations corresponding to predicates in $\vect{R}$, plus those corresponding to predicates in $\vect{S}$ (i.e., relations in $\vect{Q}$). 

When evaluating \relation{PHI} on the new database $[D, \vect{\Xi}]$, where $\vect{\Xi}$ are the extensions of relations in $\vect{Q}$ corresponding to the extensions of predicates in $\vect{\Sigma}$, we obtain that for all tuples $\langle \vect{X} \rangle$ there exists a tuple $\langle \vect{Y} \rangle$ such that the tuple $\langle \vect{X}, \vect{Y} \rangle$ belongs to \relation{PHI}, i.e.:
\begin{equation*}
\forall \langle \vect{X} \rangle \exists \langle \vect{Y} \rangle : \langle \vect{X}, \vect{Y} \rangle \in \relation{PHI}.
\end{equation*}

\noindent
Since $\langle \vect{X} \rangle \in \dom^{|\vect{X}|}$, we obtain that
$\dom^{|\vect{X}|} \subseteq \algop{\pi}{\vect{X}}(\relation{PHI})$, implying that the expression for \fail\ in the \npalg\ query~\eqref{eqn:theorem} evaluates to the empty relation for the extension $\vect{\Xi}$ of the guessed tables $\vect{Q}$.

\bigskip \noindent
(If part) Suppose that $D \not\in \vect{D}$. This implies that 
$%\begin{equation*}
D \models \lnot  (\exists \vect{S}) (\forall \vect{X}) (\exists \vect{Y}) \varphi (\vect{X}, \vect{Y} ) 
$ %\end{equation*}
or, equivalently, that:
\begin{equation*}
D \models (\forall \vect{S}) (\exists \vect{X}) (\forall \vect{Y}) \ \lnot \varphi (\vect{X}, \vect{Y} ) 
\end{equation*}

\noindent
By translating formula $\varphi$ into relational algebra, we obtain that, for every extension $\vect{\Xi}$ of relations in $\vect{Q}$ (corresponding to predicates in $\vect{S}$), there exists at least one tuple $\langle \vect{X} \rangle$ such that for every tuple $\langle \vect{Y} \rangle$, tuple $\langle \vect{X}, \vect{Y} \rangle$ does not belong to \relation{PHI}.
This implies that:
\begin{equation*}
\exists \langle \vect{X} \rangle \in \dom^{|\vect{X}|} : \langle \vect{X} \rangle \not\in \algop{\pi}{\vect{X}}(PHI),
\end{equation*}
and so that the expression for \fail\ in the \npalg\ query~\eqref{eqn:theorem} does not evaluate to the empty relation for all possible extensions $\vect{\Xi}$ of the guessed tables $\vect{Q}$.
\end{proof}
\end{theorem}

\subsection{Polynomial fragments}
\label{sec:polynomial}
Polynomial fragments of second-order logic have been presented in, e.g.,
\cite{gottlob-etal:04:existential}.  In this section we use some of those results to show that it is
possible to isolate polynomial fragments of \npalg.

\begin{theorem}
\label{thm:poly:Eaa}
Let $s$ be a positive integer, \relation{PHI} a $q$-free expression of relational algebra over the relational vocabulary $edb(D) \cup \{Q^{(s)}\}$,
and $Y_1,Y_2$ the names of two
attributes of \relation{PHI}.  An \npalg\ query of the form:
\begin{align*}
    &\guess\ Q^{(s)}; \\
    &\fail\ = \left( \dom \times \dom \right)
    - \algop{\pi}{Y_1,Y_2}\left(\relation{PHI}\right).
\end{align*}
can be evaluated in polynomial time in the size of the database.
\begin{proof}
This class of \npalg\ queries corresponds to the
$Eaa$ prefix class of second-order logic described in \cite{gottlob-etal:04:existential}, which is proved to be polynomial by a mapping into instances of 2SAT.
The correctness of the translation is formally guaranteed by Theorem~\ref{thm:expressiveness}.
\end{proof}
\end{theorem}

\noindent
Some interesting queries obeying the above restriction can indeed be
formulated. As an example, \emph{2-coloring} can be specified as follows
(when $k=2$, $k$-coloring, cf.~Section~\ref{sec:coloring}, becomes
polynomial):

\begin{align*}
&\guess\  C^{(1)}; \\*
&\fail  =   \dom \times \dom - \
%& \qquad\quad\ 
\left[
\begin{aligned}[c]
  & complement(\relation{EDGES}) \ \cup \\
  & C \algop{\times}{} complement(C) \ \cup \ 
  %& 
    complement(C) \algop{\times}{} C
\end{aligned}
\right].
\end{align*}
$C$ and its complement denote the 2-partition. The constraint states that each
edge must go from one subset to the other one.

Another polynomial problem of this class is \emph{2-partition into cliques}
(cf., e.g., \cite{garey-johnson:79:book}), which amounts to decide whether there is a
2-partition of the nodes of a graph such that the two induced subgraphs are
complete.  
An \npalg\ query which specifies the problem is:

\begin{align*}
  &\guess\ P^{(1)}; \\*
  &\fail =  \dom \times \dom - \\
  &\qquad \qquad \left[
    %%\begin{aligned}[c]
                                %&
    complement(P) \times P \ \cup \
                                %&
    P \times complement(P) \ \cup \
          %& 
    \relation{EDGES}
        %\end{aligned}
  \right].
\end{align*}

\noindent
A second polynomial class is defined
by the following theorem.

\begin{theorem}
\label{thm:poly:E1e*aa}
Let $\relation{PHI}(X_1,\ldots,X_k,Y_1, Y_2)~(k>0)$ be a $q$-free expression of relational algebra over the relational vocabulary $edb(D)\cup \{\relation{Q}^{(1)}\}$.
An \npalg\ query of the form:
  \begin{align*}
    &\guess\ \relation{Q}^{(1)}; \\
    & \relation{X}(X_1,\ldots,X_k) = %\\*
    %& \qquad 
    \relation{PHI}(X_1,\ldots,X_k,Y_1, Y_2) \ /\
    %& \qquad\ 
    \algop{\rho}{ 
                        \begin{subarray}{l}
                                \$1 \to Y_1 \\*
                                \$2 \to Y_2
                        \end{subarray}
                }     (\dom \times \dom); \\*
    &\fail\ = empty(\relation{X}).
  \end{align*}
can be evaluated in polynomial time in the size of the database.
\begin{proof}
This class of \npalg\ queries corresponds to the
$E_1e^*aa$ prefix class of second-order logic~\cite{gottlob-etal:04:existential}, which, in turn, is proved to be polynomial by a mapping into 2SAT.
Also in this case, the correctness of the translation is formally guaranteed by Theorem~\ref{thm:expressiveness}.
\end{proof}
\end{theorem}

\noindent
As an example, the specification for the \emph{Graph disconnectivity} problem,
i.e., to check whether a graph is not connected, belongs to this class.

\section{The \consql\ language}
\label{sec:consql}

In this section we describe the \consql\ language, a non-deterministic extension
of \sql\ (able to express also optimization problems) whose optimization-free subset has the same expressive power as \npalg, and present some
specifications written in this language.

\subsection{Syntax of \consql}
\consql\ is a strict superset of \sql. The problem instance is described as a
set of ordinary tables, using the data definition language of \sql. The novel
construct \texttt{CREATE SPECIFICATION} is used to define a problem specification. It has three
parts, two of which correspond to the parts of Definition~\ref{def:npalg-syn}: 

\begin{enumerate}
\item Definition of the guessed tables, by means of the new keyword
  \texttt{GUESS};
 \item Optional definition of an objective function, by means of one of the two keywords \texttt{MAXIMIZE} and \texttt{MINIMIZE};
\item Specification of the constraints that must be satisfied by guessed
  tables, by means of the standard \sql{} keyword \texttt{CHECK}.
\end{enumerate}

\noindent
Furthermore, the user can specify the desired output by means of the new
keyword \texttt{RETURN}. In particular, the output is computed when an
extension of the guessed tables satisfying all constraints and such that the objective function is optimized is found.  Of
course, it is possible to specify many guessed tables, constraints and returned
tables. The syntax is as follows (we write it in BNF, with terminals either capitalized or quoted, and, for every terminal or non-terminal $a$, ``$[a]$'' meaning optionality, ``$a*$'' a list of an arbitrary number of $a$, and ``$a+$'' meaning ``$a(a*)$''):

\bigskip
\begin{small}
\begin{verbatim}
  CREATE SPECIFICATION problem_name `('
         (GUESS TABLE table_name [`('aliases`)'] AS guessed_table_spec)+
         ((MAXIMIZE | MINIMIZE) `('aggregate_query`)'
         (CHECK `(' condition `)')+
         (RETURN TABLE return_table_name AS query)*
  `)'
\end{verbatim}
\end{small}
\bigskip

\noindent
The guessed table \verb+table_name+ gets its schema from its definition
\texttt{guess\-ed\_ta\-ble\_sp\-ec}. The latter expression is similar to a standard
\verb+SELECT-FROM-+\verb+WHE+\-\verb+RE+ \sql\ query, except for the \verb+FROM+ clause that
can contain also expressions such as:

\bigskip
\begin{small}
\begin{verbatim}
  SUBSET OF SQL_from_clause |
  [TOTAL | PARTIAL] FUNCTION_TO `(' (range_table | min `..' max) `)' 
        AS field_name_list OF SQL_from_clause |
  (PARTITION `(' n `)' | PERMUTATION) AS field_name OF SQL_from_clause
\end{verbatim}
\end{small}
\bigskip

\noindent
with \verb+SQL_from_clause+ being the content of an ordinary \sql\ \texttt{FROM} clause (e.g., a list of tables).
The schema of such expressions consists in the attributes of \verb+SQL_from_clause+,
plus the extra \verb+field_name+ (or \verb+field_name_list+), if present.

In the \verb+FROM+ clause the user is supposed to specify the shape of the
search space, either as a plain subset (like in \npalg), or as a mapping (i.e.,
partition, permutation, or function) from the domain defined by
\verb+SQL_from_clause+.  
Mappings require the specification of the range and the name of the extra
field(s) containing range values.  As for \verb+PERMUTATION+, the range is
implicitly defined to be a subset of integers.  As for \verb+FUNCTION_TO+ the
range can be either an interval \verb+min..max+ of a \sql\ enumerable type,
(e.g., integers) or the set of values of the primary key of a table denoted by
\verb+range_table+. The optional keyword \verb+PARTIAL+ means that the function
can be defined over a subset of the domain (the default is \verb+TOTAL+).  We
remind the reader that using partitions, permutations or functions does not add any
expressive power to the language 
(cf.\ Section~\ref{sec:syntsugar}).

As for the objective function, the user is supposed to specify a query whose output is a monadic table with only one tuple of an \sql\ totally ordered type (e.g., integers or reals), typically by making use of \sql\ aggregate operators like \texttt{COUNT}, \texttt{SUM}, etc.

It is possible to specify constraints on the guessed tables by using ordinary \sql\ boolean conditions, e.g., \texttt{EXISTS}, \texttt{NOT EXISTS}, \texttt{IN}, \texttt{NOT IN}, \texttt{=ANY}, \texttt{=ALL}, etc.

Finally, the \verb+query+ that defines a returned table is an ordinary \sql\ query on the tables defining the problem instance plus the guessed ones, and it is evaluated for an arbitrary extension of the guessed tables encoding an optimal solution. This is consistent with the semantics adopted by all state-of-the-art systems for Constraint Programming.

Once a problem has been specified, its solution can be obtained with an ordinary \sql\ query on the return tables: 

\bigskip
\begin{small}
\begin{verbatim}
  SELECT field_name_list 
  FROM problem_name.return_table_name 
  WHERE condition
\end{verbatim}
\end{small}
\bigskip

\noindent
The table \texttt{ANSWER(n INTEGER)} is implicitly defined locally to the
\texttt{CREATE SPEC\-IF\-ICAT\-ION} construct, and it is empty if and only if the problem has no
solution.

\subsection{Examples}
\label{sec:npalg-consql:sqlexamples}
\label{sec:sqlexamples}
In this subsection we exhibit the specification of some problems in \consql. In
particular, to highlight its similarity with \npalg, we show the specification
of the graph coloring problem of 
of Section~\ref{sec:coloring}.
Afterwards, we exploit the full power of the language and show how some
real-world problems can be easily specified. In all the examples, we describe the schema of the input database, and underline key fields.

\subsubsection{Graph $k$-coloring}
We assume an input database over the schema shown in the Entity-Relationship (ER) diagram in Figure~\ref{fig:consql:er:coloring}, thus containing relations \relation{NODES(\underline{n})},
\relation{EDGES(\underline{f},\underline{t})} (encoding the graph), and \relation{COLORS(\underline{id},name)} (listing
the $k$ colors). 
Once a database (i.e., a problem instance) has been created (by using standard \sql\ commands), a \consql\ specification of the $k$-coloring problem is the following:

\myfigure{fig:consql:er:coloring}{ER diagram of the database schema for the $k$-coloring problem. The guessed table \relation{COLORING} is in boldface.}{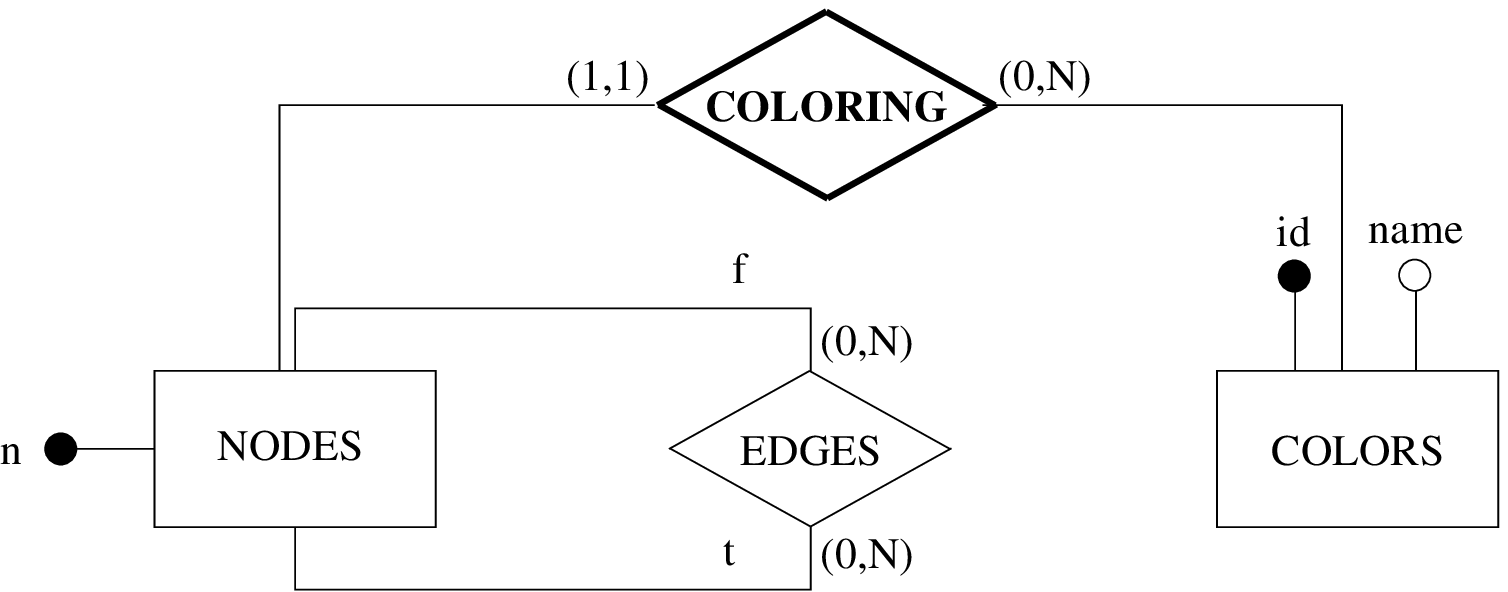}

\bigskip
\begin{small}
\begin{verbatim}
  CREATE SPECIFICATION Graph_Coloring ( 
     /* COLORING contains tuples of the kind <NODES.n, COLORS.id>, 
        with COLORS.id arbitrarily chosen. */
   GUESS TABLE COLORING AS              
     SELECT n, color FROM TOTAL FUNCTION_TO(COLORS) AS color OF NODES
      CHECK ( NOT EXISTS ( 
        SELECT * FROM COLORING C1, COLORING C2, EDGES
        WHERE C1.n <> C2.n AND C1.color = C2.color 
          AND C1.n = EDGES.f AND C2.n = EDGES.t ))
      RETURN TABLE SOLUTION AS SELECT COLORING.n, COLORS.name
        FROM COLORING, COLORS WHERE COLORING.color = COLORS.id
  )
\end{verbatim}
\end{small}
\bigskip

\noindent
The \verb+GUESS+ part of the problem specification defines a new (binary) table \verb+COLO+\-\verb+RING+, with fields \verb+n+ and \verb+color+, as a total function from the set of \verb+NODES+ to the set of \verb+COLORS+. The \verb+CHECK+ statement expresses the constraint an extension of \verb+COLORING+ table must satisfy to be a solution to the problem, i.e., there are no two distinct nodes linked by an edge which are assigned the same color.

The \verb+RETURN+ statement defines the output of the problem by a query that is evaluated for an extension of the guessed table that satisfies every constraint. The user can ask for such a solution with the statement 

\bigskip
\begin{small}
\begin{verbatim}
  SELECT * FROM Graph_Coloring.SOLUTION
\end{verbatim}
\end{small}
\bigskip

\noindent
As described in the previous subsection, if no coloring exists, the system table
\verb+Graph_Coloring+\texttt{.AN\-SWER} will contain no tuples. This can be
easily checked by the user, in order to obtain only a significant
\verb+Graph_Coloring.SOLUTION+ table.

\subsubsection{University course timetabling}
\label{subsec:timetabling}

The \emph{University course timetabling} problem \cite{scha-99} consists in
finding the weekly scheduling for all the lectures of a set of university
courses in a given set of classrooms. We consider a variant of the original problem in which the objective function to minimize is the total number of students that have to attend overlapping lectures.

The input database schema is shown in Figure~\ref{fig:consql:er:university}, and consists of the following relations:

\myfigure{fig:consql:er:university}{ER diagram of the database schema for the University course \mbox{timetabling problem}. The guessed table \relation{TIMETABLE} is in boldface.}{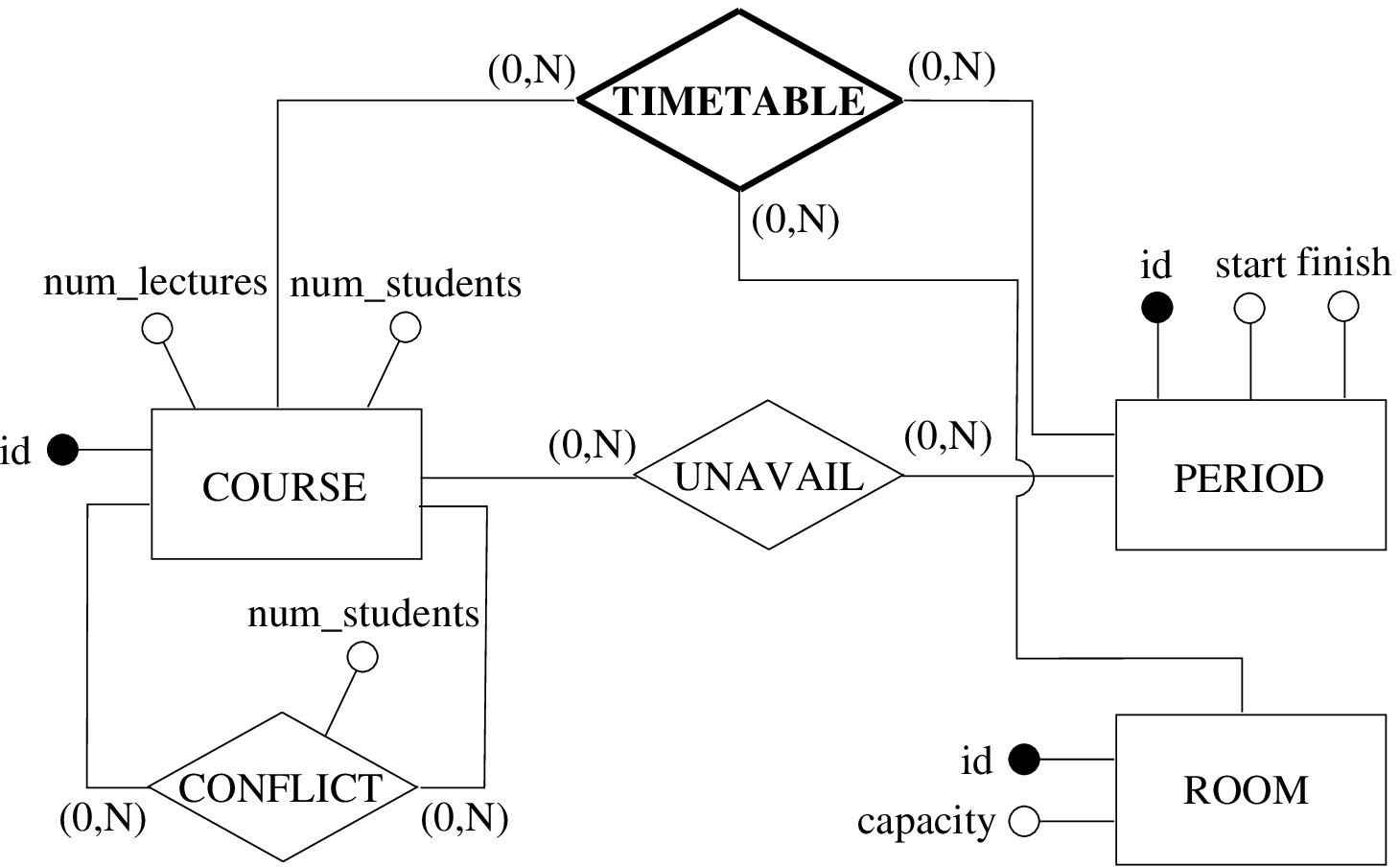}

\begin{itemize}

\item \relation{COURSE({\underline {id}}, num\_lectures, num\_students)}, consisting of
  tuples $\langle \relation{c}, \relation{l}, \relation{s} \rangle$ meaning that the course \relation{c} needs \relation{l} lectures a week, and has \relation{s} enrolled students. 

\item \relation{PERIOD({\underline {id}}, start, finish)} encoding (non-overlapping) periods,
  plus information on start and finish time.

\item \relation{ROOM({\underline {id}}, capacity)}. A tuple $\langle \relation{r}, \relation{c} \rangle$ means that room \relation{r} has capacity \relation{c}.

\item \relation{CONFLICT({\underline {course1}}, {\underline {course2}}, num\_students)}. A tuple
  $\langle \relation{c1}, \relation{c2}, \relation{n} \rangle$ means that courses \relation{c1} and \relation{c2} have \relation{n} common
  students.

\item \relation{UNAVAIL({\underline {course}}, {\underline {period}})}. A tuple
  $\langle \relation{c}, \relation{p} \rangle$ means that the teacher of course \relation{c} is not available
  for teaching at period \relation{p}. 

\end{itemize}

\noindent
A solution to the problem is a (guessed) relation \relation{TIMETABLE(period, room,
course)} with tuples $\langle \relation{p}, \relation{r}, \relation{c} \rangle$ meaning that at period \relation{p} in
room \relation{r} there is a lecture of course \relation{c}. If for some values of the \relation{room} and
\relation{period} fields there is no tuple in the relation \relation{TIMETABLE}, then the room is unused in that period.

A \consql\ specification of the timetabling problem, given an input database,
is the following:

\bigskip
\begin{small}
\begin{verbatim}
  CREATE SPECIFICATION University_Timetabling (
    GUESS TABLE TIMETABLE(period, room, course) AS
      SELECT p.id, r.id, course
      FROM PARTIAL FUNCTION_TO(COURSE) AS course OF PERIOD p, ROOM r
    // Objective function
    MINIMIZE ( SELECT SUM(c.num_students)
      FROM TIMETABLE t1, TIMETABLE t2, CONFLICT c
      WHERE t1.period = t2.period AND t1.course <> t2.course AND
            c.course1 = t1.course AND c.course2 = t2.course
    )  
    // At most one lecture of a course per period
    CHECK ( NOT EXISTS (
      SELECT * FROM TIMETABLE t1, TIMETABLE t2
      WHERE t1.course = t2.course AND
            t1.period = t2.period AND t1.room <> t2.room
    ))
    // Unavailability constraints
    CHECK ( NOT EXISTS (
      SELECT * FROM TIMETABLE t, UNAVAIL u
      WHERE t.course = u.course AND t.period = u.period
    ))
    // Capacity constraints
    CHECK ( NOT EXISTS (
      SELECT * FROM TIMETABLE t, COURSE c, ROOM r
      WHERE t.course = c.id AND t.room = r.id AND
            c.num_students > r.capacity
    ))
    // Teaching requirements
    CHECK ( NOT EXISTS (
      SELECT * FROM COURSE c
      WHERE c.num_lectures <> 
        ( SELECT COUNT(*) FROM TIMETABLE t
            WHERE t.course = c.id
        )
    ))
    RETURN TABLE SOLUTION AS SELECT * FROM TIMETABLE
  )
\end{verbatim}
\end{small}

\bigskip
\noindent
In particular, the constraints force extensions of the guessed table \relation{TIMETABLE} to be such that:
\begin{itemize}
\item There is at most one lecture of a course per period, i.e., there cannot be two different rooms allocated for the same course in the same time slot;
\item Unavailability constraints are respected, i.e., no lecture is scheduled in a period for which the relevant teacher is unavailable;
\item Capacity constraints are respected, i.e., no room is allocated for courses having a number of students that exceeds its capacity;
\item Teaching requirements are satisfied, i.e., all courses have a room and a time slot assigned for all the lectures they need.
\end{itemize}

\noindent
An extension for guessed table \relation{TIMETABLE} that satisfies the constraints above is an optimal solution to the University course timetabling problem if it minimizes the overall number of students that are expected to attend conflicting lectures, i.e., lectures that are scheduled at the same time.

\subsubsection{Aircraft landing}
\label{subsec:aircraft}

The \emph{aircraft landing} problem \cite{beas-etal-00} consists in scheduling
landing times for aircraft. Upon entering within the radar range of the air
traffic control (ATC) at an airport, a plane requires a \emph{landing time} and
a \emph{runway} on which to land. The landing time must lie within a specified
time window, bounded by an \emph{earliest time} and a \emph{latest time},
depending on the kind of the aircraft.
Each plane has a most economical, preferred speed. A plane is said to be
assigned its \emph{target time}, if it is required to fly in to land at its
preferred speed. If ATC requires the plane to either slow down
or speed up, a cost incurs. The bigger the difference
between the assigned landing time and the target landing time, the bigger the
cost.
Moreover, the amount of time between two landings
must be greater than a specified minimum (the
\emph{separation time}) that depends on the planes involved.
Separation times depend on the aircraft landing on the same
or different runways (in the latter case they are smaller).

Our objective is to find a landing time for each planned aircraft, encoded in a
guessed relation $\relation{LANDING}$, satisfying all the previous constraints, and such
that the total cost (i.e., the sum of the costs associated with each aircraft) is minimized.
The input database schema is shown in Figure~\ref{fig:consql:er:aircraft}, and consists of the following relations:

\myfigure{fig:consql:er:aircraft}{ER diagram of the database schema for the Aircraft landing problem. The guessed table \relation{LANDING} is in boldface.}{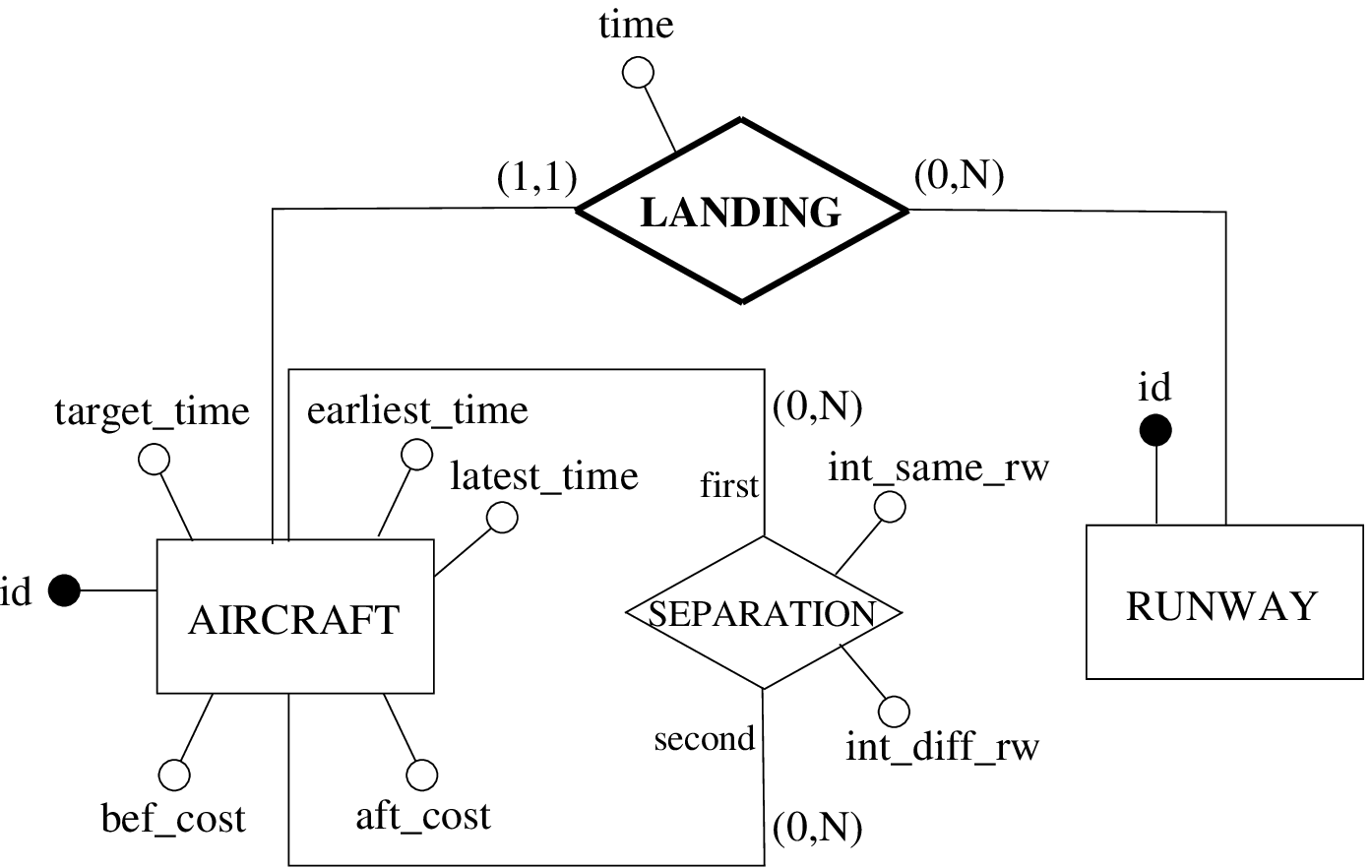}

\begin{itemize}
\item $\relation{AIRCRAFT(\underline {id}, target\_time, earliest\_time, latest\_time, bef\_cost, aft\_cost)}$,
  listing aircraft planned to land, together with their target times and
  landing time windows; the cost associated with a delayed or advanced landing
  at time \relation{x} is given by $\relation{bef\_cost} \cdot \max[0, \relation{t}-\relation{x}] \ + \ \relation{aft\_cost} \cdot
  \max[0, \relation{x}-\relation{t}]$, where $\relation{t}$ is the aircraft target time.

\item $\relation{RUNWAY(\underline {id})}$ listing all the runways of the airport.

\item $\relation{SEPARATION(\underline {first}, \underline {second}, int\_same\_rw, int\_diff\_rw)}$.  
  A tuple $\langle \relation{a}, \relation{a'}, \relation{is}, \relation{id} \rangle$ means that if aircraft $\relation{a'}$ lands after
  aircraft $\relation{a}$, then landing times must be separated by $\relation{is}$ (resp.\ $\relation{id}$) minutes if they land
  on the same runway (resp.\ on different runways).

\end{itemize}

\noindent
In the following specification, the search space is a total function 
assigning an aircraft to a landing time and a runway. For the sake of simplicity, landing times are expressed in minutes after a conventional time instant, e.g., the scheduling starting time, and the time horizon is set to one day, i.e., $24 \times 60$ minutes.

\bigskip
\begin{small}
\begin{verbatim}
  CREATE SPECIFICATION Aircraft_Landing (
    GUESS TABLE LANDING AS
      SELECT ar.id AS aircraft, ar.runway, at.time
      FROM (TOTAL FUNCTION_TO(RUNWAY) AS runway OF AIRCRAFT) ar,
           (TOTAL FUNCTION_TO(0..24*60-1) AS time OF AIRCRAFT) at
      WHERE ar.id = at.id
     // Objective function
     MINIMIZE ( SELECT SUM(cost) 
       FROM ( 
         SELECT a.id, (a.bef_cost * (a.target_time - l.time)) AS cost
           FROM AIRCRAFT a, LANDING l 
           WHERE a.id = l.aircraft AND l.time <= a.target_time 
         UNION // advanced plus delayed aircraft
         SELECT a.id, (a.aft_cost * (l.time - a.target_time)) AS cost
           FROM AIRCRAFT a, LANDING l 
           WHERE a.id = l.aircraft AND l.time > a.target_time 
       ) AIRCRAFT_COST // Contains tuples <aircraft, cost>
     )
     // Time window constraints
     CHECK ( NOT EXISTS ( 
       SELECT * FROM LANDING l, AIRCRAFT a WHERE l.aircraft = a.id 
         AND ( l.time > a.latest_time OR l.time < a.earliest_time ) 
     ))
     // Separation constraints
     CHECK ( NOT EXISTS ( 
       SELECT * FROM LANDING l1, LANDING l2, SEPARATION sep
       WHERE l1.aircraft <> l2.aircraft AND l1.time <= l2.time AND 
         sep.first = l1.aircraft AND sep.second = l2.aircraft AND 
         (  ( (l1.runway = l2.runway) AND 
              (l2.time - l1.time) < sep.int_same_rw ) OR
            ( (l1.runway <> l2.runway) AND 
              (l2.time - l1.time) < sep.int_diff_rw )
     ))) 
     RETURN TABLE SOLUTION AS SELECT * FROM LANDING
  )
\end{verbatim}
\end{small}

\bigskip
\noindent
In particular, the constraints force extensions of the guessed table \relation{LANDING} to be such to respect both time window constraints (i.e., the actual landing time for each aircraft must lie inside its landing time window), and separation constraints (encoded in the \relation{SEPARATION} relation). Such an extension is an optimal solution to the Aircraft landing problem if it minimizes the overall cost.

\subsection{\consqlsim}
\label{subsec:consqlsimulator}
\begin{figure}[tbp]
%\psfrag{Query}{\scriptsize{Query}}
\psfrag{JLocal}{\small{\jlocal}}
\psfrag{conSql simulator}{\scriptsize{\consqlsim}}
%\psfrag{R-DBMS}{\scriptsize{R-DBMS}}
%\psfrag{JDBC}{\scriptsize{JDBC}}
%\bigskip
\begin{center}
\includegraphics[width=0.6\textwidth]{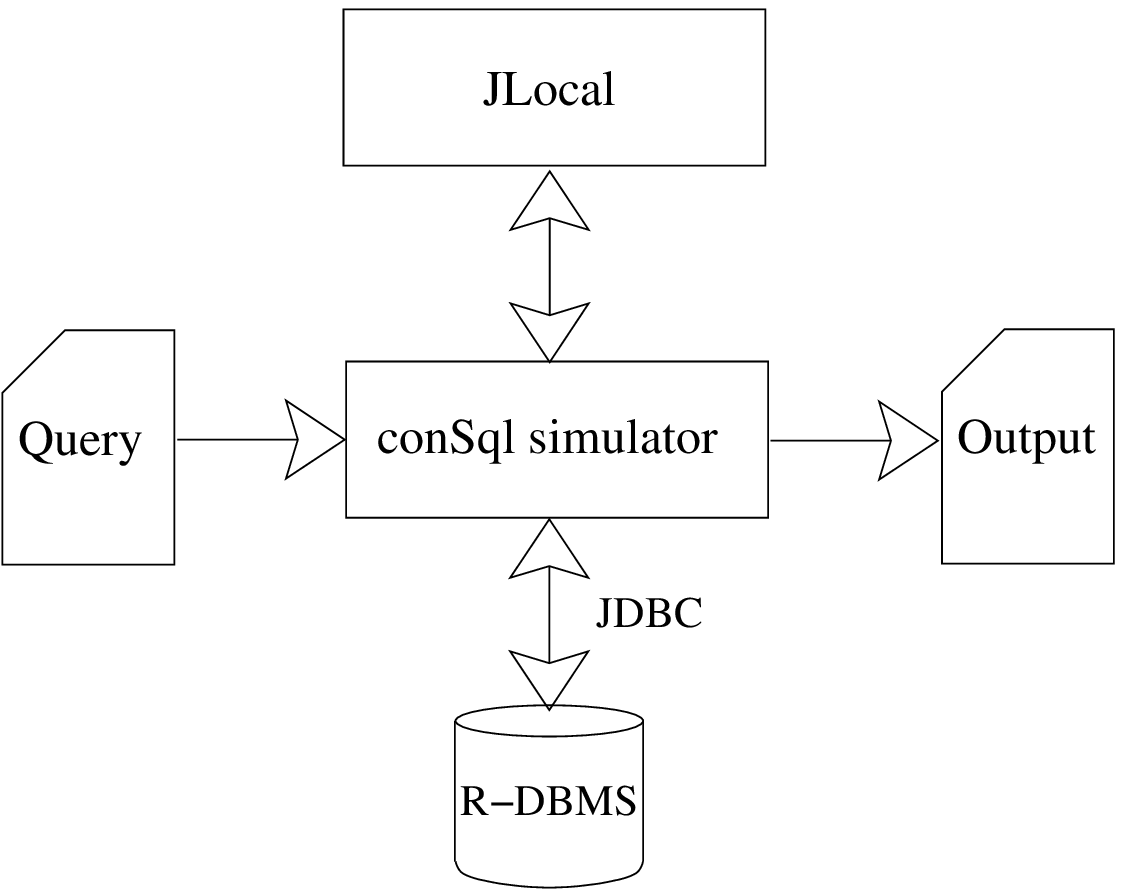}
\caption{\label{fig:consqlsim:architecture}\consqlsim\ overall architecture}
\end{center}
\end{figure}

\consqlsim\ is an application that works as an interface to a
traditional R-DBMS. It simulates the behavior of a \consql\ server by reading from its input stream
\consql\ queries, i.e., ordinary \sql\ queries and commands, and problem specifications. Ordinary \sql\ queries and commands are simply passed to the underlying R-DBMS, while problem specifications are processed. 
The overall architecture of the system is depicted in Figure~\ref{fig:consqlsim:architecture}.
In particular, \texttt{CREATE SPECIFICATION} constructs are parsed, creating the new tables
(corresponding to the guessed ones) and an internal representation of the
search space. 
The search space is then explored by the solver, looking for an element corresponding to an optimal
solution, by posing appropriate queries to the R-DBMS (in standard \sql).  
As soon as an optimal solution is found, results of the queries 
specified in the \texttt{RETURN} statements are accessible to the user as output.

\begin{figure}[tbp]
\psfrag{Problem}{\scriptsize{\textit{Problem}}}
\psfrag{Constraint}{\scriptsize{\textit{Constraint}}}
\psfrag{SearchSpace}{\scriptsize{\textit{SearchSpace}}}
\psfrag{ObjFunction}{\scriptsize{\textit{ObjFunction}}}
%\bigskip
\begin{center}
\includegraphics[width=0.8\textwidth]{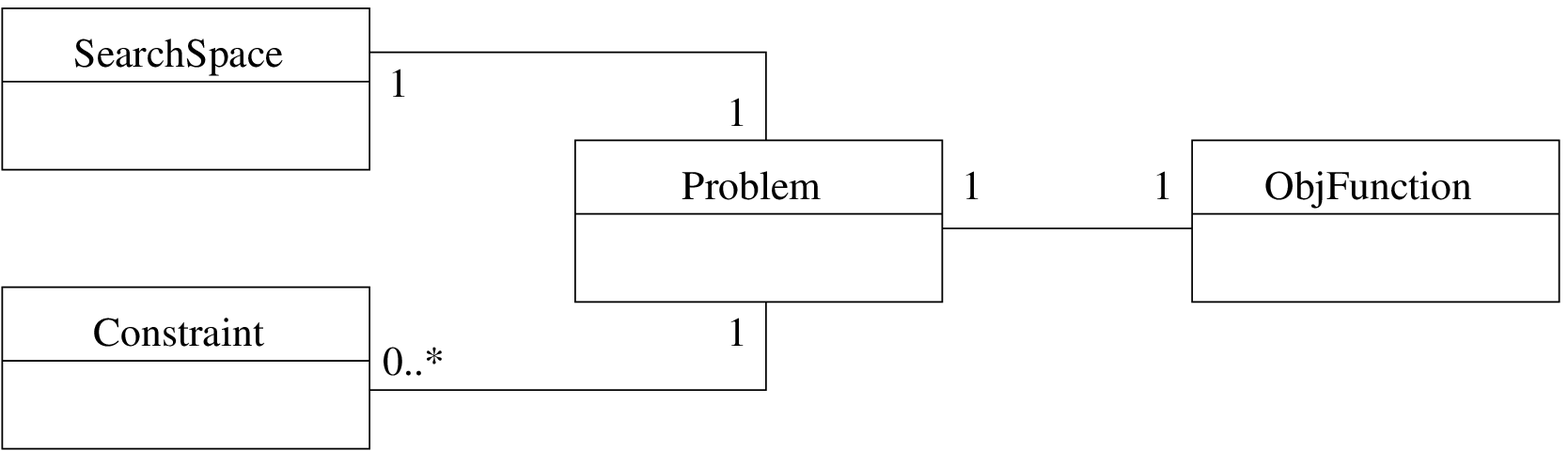}
\caption{\label{fig:consqlsim:uml:problem}Portion of the conceptual UML class diagram for the language-independent problem modelling module.}
\end{center}
\end{figure}

The implementation of \consqlsim\  gives much attention to software engineering aspects and to different quality factors of software artifacts.
In particular, the system is platform independent and highly portable, since it is written in Java, and uses the standard JDBC protocol for the connection with the R-DBMS, and the whole architecture presents a neat separation among the language parser, the problem modelling module and the solver engine (\jlocal), 
so as to emphasize qualities such as modularity, extendability and reusability. In particular, the problem modelling module allows to represent problem specifications in a \emph{language independent} fashion, relying on abstract concepts such as \emph{Problem}, \emph{Search space}, \emph{Objective function}, and \emph{Constraint}, as the conceptual UML diagram in Figure~\ref{fig:consqlsim:uml:problem} shows. 
In this way, the solving engine \jlocal, interacting with the abstract problem modelling module, is independent of the particular language, i.e., \consql.
The only language-dependent part of the system is the parser for the \consql\ language, that provides concrete implementations for the abstract concepts that compose the problem modelling module, building the internal representation of the problem instance, the search space, and the constraints, and for some of the services needed by the solver.

As for the search methods, the solver engine \jlocal\ exploits local search techniques to find solutions. 
Local search is considered one of the most attractive techniques for solving combinatorial optimization problems (cf., e.g., \cite{aarts-lenstra:97:local}), being able to solve instances of realistic size in reasonable time.
Several local search algorithms have been implemented in \jlocal, among them Hill climbing and Tabu search \cite{glov-lagu-97}, and several strategies that combine different solvers for doing the search are present (e.g., Tandem search, in which two different solvers are used in sequence). 
Additional local search strategies can be simply added by subclassing the \emph{LocalSearchSolver} class (not described here for the sake of simplicity). 

It is worth noting that the user is completely unaware of the search techniques implemented by the system. In particular, definitions for neighborhoods, moves, aspiration functions, etc., are made by \consqlsim\ itself, starting from the types of the guessed tables defined in the specification (i.e., subsets, functions, permutations, partitions).

The development of \consqlsim\ has been done according to the iterative model of the software life-cycle.
In particular, three iterations were expected. The first, prototypical, version of the system, which was used only to test specifications, 
relied on a purely enumerative approach, and, of course, no considerations on performances could be done.
As for the present version, which is at the second iteration of the development process, we added the local search engine, but the connection with the DBMS is yet completely black box. In particular, \jlocal\ uses the DBMS both for maintaining the current state, for checking constraints, and for evaluating which neighbor to visit next.
The main motivation behind this choice, is that current DBMSs offer means to answer queries efficiently, especially in case of very large instances.
Nonetheless, since constraints are evaluated from scratch in every visited state, performances cannot be good, and only instances of small sizes can be actually solved.

In the third version of the system, which is currently under development, we are adding the following functionalities:
\begin{enumerate}
\item The ability of checking constraints incrementally: constraints' check is the main source of inefficiency of the current version, since they are evaluated from scratch for every visited state, and for all its neighbors, in order to choose the best move. Hence, the number of queries posted to the DBMS is very high, and all of them are answered independently from each other.
However, it is clear that, when using local search technology, only a small variation in the number of constraint violations is expected, when moving from one state to its neighbors. To this end, our goal is to make the DBMS able to compute only variations to constraints' violations when performing checks. This is expected to greatly increase the overall performances of the system, since we can rely on very sophisticated algorithms to, e.g., maintain and synchronize views, actually present in currently available DBMSs.

\item The use of a much more complex local search engine. In particular, we are currently integrating EasyLocal++~\cite{digaspero-schaerf:03:easylocal}, a very sophisticated solver, with our system. This can lead to better algorithms, and to a fine tuning of their parameters.

\item The addition of an optional ``search'' part in \texttt{CREATE SPECIFICATION} constructs, as it happens in, e.g., \opl, in order to provide the user with the possibility of declaring which search algorithm to adopt, 
as well as the types of neighborhoods, moves, aspiration functions, etc.
Of course, as already claimed, our goal is to provide good defaults for the options in this part, as, e.g., \opl\ does, by letting the system able to automatically make a good choice for these issues, depending on the specification at hand.

\item To provide a better coupling with a particular open-source DBMS, in order to make the system able to directly use the DBMS' APIs, instead of interacting by means of (inefficient, but highly portable) protocols like JDBC.
\end{enumerate}

\section{Conclusions, related and future work}
\label{sec:related}

In this paper we have tackled the issue of strong integration between
constraint modelling and programming and up-to-date technology for storing data. In
particular we have proposed constraint languages which have the ability to
interact with data repositories in a standard way.  To this end, we have
presented \npalg, an extension of relational algebra which is specially suited
for combinatorial problems.  The main feature of \npalg\ is the possibility of
specifying, via a form of non-determinism, a set of relations that can have an
arbitrary extension.  This allows the specification of a search space suitable
for the solution of combinatorial problems, with ordinary relational algebra expressions
defining constraints.  Although \npalg\ provides just a very simple guessing
operator, many useful search spaces, e.g., permutations and functions, can be
defined as syntactic sugar.

Several computational properties of \npalg\ have been shown, including data and
combined complexity, and expressive power. Notably, the language is shown to
capture exactly all the problems in the complexity class NP, which includes
many combinatorial problems of industrial relevance.
In the same way, we have proposed \consql, a non-deterministic extension of
\sql, with the same expressive power of \npalg, which is suitable also for specifying optimization problems.  The effectiveness of \npalg\ 
and \consql\ both as complex query and constraint modelling languages has been
demonstrated by showing several queries which specify combinatorial problems.

Other extensions of relational algebra have already been proposed. The most important examples are the languages $\relation{Alg+while}$ and $\relation{Alg+while^+}$, where, respectively, a non-inflationary and an inflationary fixpoint semantics is added~\cite{abit-hull-vian-95}. 
Both these languages are capable of expressing the Transitive Closure query, but have very different expressive power: $\relation{Alg+while}$ can express queries in PSPACE, but the language captures exactly this class only on ordered databases (i.e., databases in which a total order among all constants occurring in it is fixed).
As for $\relation{Alg+while^+}$ instead, it can express only polynomial-time queries, and the language captures the whole PTIME class only on ordered databases.
A feature for expressing linear recursion has recently been added also to \sql\ (\sql`99), by means of the \texttt{WITH} construct.
However, both the aforementioned extensions of relational algebra, and the new version of \sql\ do not make such languages suitable for expressing constraint problems.

Several languages and systems for constraint programming are nowadays available either as research and commercial packages. 
Some of them are in the form of frameworks and libraries. 
As an example, in $ECL^iPS^e$~\cite{eclipse} or \textsc{SICStus}~\cite{sicstus} a traditional programming language such as \textsc{Prolog} is enhanced by means of libraries and specific constructs for specifying constraints, which are then solved by highly optimized algorithms. 
The ILOG Optimization suite \cite{ILOG98} provides instead libraries for expressing constraints callable by host general-purpose programming languages like C++.

Specification languages natively developed for constraint modelling and programming are also available, either commercially like \opl\ \cite{vanh-99} and \ampl\ \cite{four-gay-kern-93} or as research prototypes, like \textsc{esra} \cite{flener-etal:04:esra}, all of them offering an ad-hoc syntax for problem specifications. 
Similarly to \npalg\ and \consql, they support a clear distinction between the
data and the problem description level, but differently from them, \npalg\ and \consql\ use standard and well-known languages for specifying problem specifications, that are considered just like queries over a relational database representing the input instance. 
We believe that this feature allows for a wider diffusion of the declarative constraint modelling paradigm in industrial environments, permitting a very strong integration with the information system of the enterprise.
Conversely, the other systems usually get input data from text files in ad-hoc formats, and additional machinery is needed to build such files from the content of a relational database, and for storing problem solutions. Even if some of them have plug-ins that can be used to make connections to databases, e.g.,~\cite{ILOG-DBLINK}, data are always processed outside the DBMS, hence leading to a potential lack of data integrity.

Several query languages capable of capturing the complexity class NP have been
shown in the literature. As an example, in \cite{kola-papa-91} an
extension of datalog (the well-known recursive query language \cite{ullman:88:principles-vol1})
allowing negation is proved to have such a property.  Another extension of
datalog capturing NP, without negation but with a form of non-determinism, is proposed in
\cite{cado-palo-98}.  
Other rule-based languages with different semantics have also been proposed: \textsc{Smodels}~\cite{simons-etal:02:stable} which relies on stable models semantics, and \textsc{dlv}~\cite{leone-etal:dlv} which is based on answer set programming.
They also are based on negation and recursion.
On the other hand, \npalg\ captures NP without recursion.
Actually, recursion can be simulated by non-determinism, and it is possible to
write, e.g., the transitive closure query in \npalg.
Being non-recursive, \npalg\ is more similar to plain existential second order 
logic. Nevertheless, it retains the functional character of relational algebra,
which sometimes makes it easier (with respect to rule-based languages) to specify a problem.

For what concerns \consql, we believe it is a clear step towards a language for both declarative constraint modelling and complex queries to relational databases, which relies on standard and well-known technologies. 
Currently, the most adopted solution for evaluating complex queries over a relational database is to embed \sql\ into a general-purpose programming language, like Java or C++, thus by processing stored data and intermediate results outside the database. \consql\ instead has been designed for being implemented \emph{inside} the DBMS, so guaranteeing all transactional properties to the query evaluation process.

As for the proposed implementation of \consqlsim, it is conceived to be based on a purely declarative language and to be ready to use, i.e., it does not require any additional code to be written by the user. 
Other systems for local search do, however, exist, either in forms of declarative languages for modelling in a concise way local search algorithms (cf., e.g., \cite{michel-vanhentenryck:00:localizer,vanhentenryck-michel:03:control}) or, alternatively, in forms of libraries or frameworks (cf., e.g., Local++ \cite{scha-cado-lenz-00}), hence providing algorithms that rely on additional application-specific code provided by the user. 
\consqlsim\ is different from such systems in that it provides the user with the ability of modelling an optimization problem by means of a language, i.e., \consql, that is completely unaware of the particular solving technology used. 
It is responsibility of the engine to provide the local search solver with all the information needed to explore the search space (e.g., description of neighborhoods, moves, etc.). This choice is currently made starting from the types of the guessed tables defined in the specification, and future work has to be done in order to better exploit the different alternatives, as discusses at the end of Subsection~\ref{subsec:consqlsimulator}.

\consqlsim\ will be released as free and potentially open source software, thus allowing the system to receive improvements and extensions from the community.

\appendix

\section{Combined complexity of \npalg}	\label{sec:proofCombined}

In this section we prove Theorem~\ref{thm:combinedcompl}.
The proof consists in reducing an NE-complete problem, \emph{Succint $3$-coloring} \cite{kola-papa-91}, i.e., the ``succinct version'' of the graph $3$-coloring problem, into an \npalg\ query. 
It is worth noting that the resulting \npalg\ query is \emph{not uniform} with respect to the problem instance, but this is exactly what the definition of \emph{combined complexity} (as opposed to \emph{data complexity}) states.
The Succint $3$-coloring problem is defined as follows:

\begin{definition}[The Succint $3$-coloring problem]
Nodes of the input graph are elements of $\{0,1\}^n$, and, instead of an explicitly given \relation{EDGES} relation, there is a \emph{boolean circuit} with $2n$ inputs and one output, such that the value output by the circuit is 1 if and only if the inputs are two $n$-tuples that encode a pair of nodes connected by an edge.
A boolean circuit is a finite set of triples $\{(a_i, b_i, c_i), i = 1,\ldots, k\}$, where $a_i \in \{OR, AND, NOT, IN\}$ is the kind of the gate, and $b_i, c_i < i$ are the inputs of the gate (hence, the whole circuit is acyclic), unless the gate is an input gate ($a_i = IN$), in which case, say, $b_i = c_i = 0$. For NOT gates, $b_i = c_i$.
Given values in $\{0, 1\}$ for the input gates, we can compute the values of all gates one by one by starting from the first one.
The value of the circuit is the value of the last gate.
Finally, the \emph{Succint $3$-coloring problem} is the following: Given a boolean circuit
with $2n$ inputs and one output, is the graph thus presented $3$-colorable?
\end{definition}

\noindent
The Succint $3$-coloring problem is proven to be NE-complete in the same paper \cite{kola-papa-91}.

\paragraph{Reduction of Succint $3$-coloring into an \npalg\ query.}
Given an input boolean circuit $G=\{g_i=(a_i, b_i, c_i)\ | \ 1 \leq i \leq k\}$ with $2n$ inputs and one output, we construct the \npalg\ query $\psi$ that specifies the Succint $3$-coloring problem (on the graph represented by circuit $G$) as follows.

As for the set \vect{Q} of guessed relations, we declare a relation $\relation{G}_i^{(2n)}$ for every gate $i$, i.e., for every triple $g_i=(a_i, b_i, c_i)$, ($1 \leq i \leq k$). Moreover, we declare in \vect{Q} three more relations, $\relation{COL}_1^{(n)}$, $\relation{COL}_2^{(n)}$, $\relation{COL}_3^{(n)}$, encoding the partition of the nodes into 3 groups, analogously to the specification for $k$-coloring given in Section~\ref{sec:coloring}.
So, the \guess\ part of the \npalg\ query being built is the following:
\begin{equation*}
\guess\ \relation{G}_1^{(2n)}, \dots, \relation{G}_k^{(2n)}, \relation{COL}_1^{(n)}, \relation{COL}_2^{(n)}, \relation{COL}_3^{(n)};
\end{equation*}

\noindent
Intuitively, relations $\relation{G}_i$ will contain all tuples $\langle \vect{X}, \vect{Y} \rangle$, with $\vect{X} = \langle X_1, \ldots, X_n \rangle$, and $\vect{Y} = \langle Y_1, \ldots, Y_n \rangle$ (i.e., binary encodings of the nodes $X$ and $Y$) for all pairs of nodes $X$ and $Y$ that make the output of the $i$-th gate 1.

The expression for \fail\ is of the following kind:
\begin{equation*}
\fail \ = \fail\relation{\_CIRCUIT} \ \cup \ \fail\relation{\_PARTITION} \ \cup \ \fail\relation{\_COLORING}.
\end{equation*}

\noindent
The first subexpression evaluates to the empty relation if and only if the guessed extension for the $\relation{G}_i$ relations correctly encodes the circuit, while the second and the third ones evaluate to the empty relation if and only if relations $\relation{COL}_1$, $\relation{COL}_2$, $\relation{COL}_3$, are a partition of the graph nodes and a correct coloring of the graph (we omit their definitions, since they are very similar to those presented in Section~\ref{sec:coloring}).

The expression for $\fail\relation{\_CIRCUIT}$ contains in turn one of the following subexpressions $\fail\_\relation{G}_i$, $1 \leq i \leq k$, for every gate $i$, according to its type $a_i$. In particular:
\begin{itemize}
\item If $a_i = AND$, then $\fail\_\relation{G}_i = G_i\ \Delta \ \left[ G_{b_i}\ \algop{\cap}{}\ G_{c_i} \right]$;

\item If $a_i = OR$, then $\fail\_\relation{G}_i = G_i\ \Delta \ \left[ G_{b_i}\ \algop{\cup}{}\ G_{c_i} \right]$;

\item If $a_i = NOT$, then $\fail\_\relation{G}_i = G_i\ \Delta \ \left[ \dom_{01}^{2n} - G_{b_i} \right]$;

\item If $a_i = IN$, then $\fail\_\relation{G}_i = G_i\ \Delta \ \algop{\sigma}{\$j=1}(\dom_{01}^{2n})$, assuming that the $i$-th gate (of type $IN$) is the $j$-th input of the circuit.
\end{itemize}

\noindent
In the above definition, we used the relation $\dom_{01}$, defined as:
\begin{equation*}
\dom_{01} = \algop{\sigma}{\begin{subarray}{c}
				\$1 \neq AND \ \land \\
				\$1 \neq OR \ \land \\
				\$1 \neq NOT \ \land \\
				\$1 \neq IN\\
			  \end{subarray}
	   } (\dom)
\end{equation*}
that will contain at most the two tuples $\langle0\rangle$ and $\langle1\rangle$ (since $\dom$ would also contain constants for the gate types).
Thus, the expression for $\fail\relation{\_CIRCUIT}$ is the following:
\begin{equation*}
\fail\relation{\_CIRCUIT} \ = \ \bigcup_{i=1}^k \ \fail\relation{\_G}_i.
\end{equation*}

\noindent
It remains to prove that the expression for $\fail\relation{\_CIRCUIT}$ evaluates to the empty relation if and only if guessed relations $\relation{G}_1, \ldots, \relation{G}_k$ correctly encode the boolean circuit representing the input graph, i.e., if and only if for all $i$, relation $\relation{G}_i$ contains exactly all $2n$-tuples (encoding pairs of nodes given as input to the circuit) that make the output of the $i$-th gate 1. This is what the following lemma claims.

\begin{lemma} \label{th:succcirc-NPAlg}
Let $G=\{g_i=(a_i, b_i, c_i)\ | \ 1 \leq i \leq k\}$ be a boolean circuit encoding a graph, and let $\psi$ be the \npalg\ query built as described above.
An extension for guessed tables $\relation{G}_1^{(2n)},\dots,\relation{G}_k^{(2n)}$ exists such that the expression for $\fail\relation{\_CIRCUIT}$ evaluates to the empty relation. Moreover, for such an extension, each $\relation{G}_i$ contains exactly all $2n$-tuples $\langle X_1,\ldots,X_n,Y_1, \ldots, Y_n\rangle$ that make the output of the $i$-th gate 1. As a consequence, the extension for $\relation{G}_k$ contains all $2n$-tuples that encode pairs of nodes linked by an edge.

\begin{proof}
We first show that, if extensions for $\relation{G}_1, \ldots \relation{G}_k$ in the \npalg\ query $\psi$ exist that make the expression for $\fail\relation{\_CIRCUIT}$ evaluate to the empty relation (by making all the expressions for $\fail\relation{\_G}_i$ evaluate to the empty relation), then, for every input $\{X_1,\ldots,X_n,Y_1, \ldots, Y_n\}$ to the circuit, each gate $i$ ($1 \leq i \leq k$) outputs 1 if and only if the $2n$-tuple $\langle X_1,\ldots,X_n,Y_1, \ldots, Y_n\rangle$ belongs to the corresponding $\relation{G}_i$. Secondly, we show that such an extension indeed exists.
The proof of the first point is by induction on the index $i$:
\begin{description}
\item[$i=1:$] Gate $g_1$ is, by construction, of type $IN$ (i.e., $a_1 = IN$). Let us assume that $g_1$ is the $j$-th input  to the circuit, i.e., its output is 1 if and only if the $j$-th input to the circuit is 1. As it can be observed from the definition of $\fail\relation{\_G}_1$, since by hypothesis it evaluates to the empty relation, $G_1$ contains all $2n$-tuples that have 1 as the $j$-th component.

\item[$i>1:$] Let us assume that the lemma holds for all $i'$ such that $1 \leq i' < i$, and let us consider the $i$-th gate (of type $a_i \in \{ IN, AND, OR, NOT \}$) and the extension for the corresponding guessed relation $\relation{G}_i$. Since, by hypothesis, $\fail\relation{\_G}_i$ evaluates to the empty relation, it can be easily observed by its definition that:
	\begin{itemize}
	\item If $a_i=IN$, assuming that $g_i$ is the $j$-th input to the circuit, $\relation{G}_i$ contains, by construction, all $2n$-tuples that have 1 as the $j$-th component;

	\item If $a_i=AND$, it follows by induction that $\relation{G}_{b_i}$ and $\relation{G}_{c_i}$ contain exactly those tuples that make the output of, respectively, gates $g_{b_i}$ e $g_{c_i}$ 1.
	By construction, the extension for $\relation{G}_i$ contains exactly those tuples that belong to both  $\relation{G}_{b_i}$ and $\relation{G}_{c_i}$.
	
	\item If $a_i=OR$ an analogous argument holds, showing that $\relation{G}_i$ contains exactly those tuples that belong to $\relation{G}_{b_i}$ or to $\relation{G}_{c_i}$.

	\item If $a_i=NOT$, it follows by induction that $\relation{G}_{b_i}$ (in this case $b_i = c_i$) contains exactly those tuples that make the output of gate $g_{b_i}$ 1.
	By construction, the extension for $\relation{G}_i$ contains exactly those tuples in $\dom_{01}^{2n}$ that do not belong to $\relation{G}_{b_i}$.
	\end{itemize}
\end{description}

\noindent
As for the second point of the proof, it is easy to show that an extension for $\relation{G}_1, \ldots \relation{G}_k$ that makes all the expressions for $\fail\relation{\_G}_i$ evaluate to the empty relation indeed exists. The key observation is that expressions for $\fail\relation{\_G}_i$ essentially define which tuples must belong to each $\relation{G}_i$ (more precisely, each $\fail\relation{\_G}_i$ evaluates to the empty relation if and only if $\relation{G}_i$ contains exactly the tuples that belong to the relational algebra expression on the right of the ``$\Delta$'' symbol), and that the \guess\ part of the query generates all possible extensions of those relations with elements in $\dom \supset \dom_{01}$.
\end{proof}
\end{lemma}

\noindent
Lemma~\ref{th:succcirc-NPAlg} claims that an extension for $\relation{G}_1, \ldots \relation{G}_k$ in query $\psi$ that makes the expression for $\fail\relation{\_CIRCUIT}$ evaluate to the empty relation exists, and is the one that correctly models the boolean circuit representing the input graph. 
It remains to prove that the whole query $\psi$ is such that $\fail\Diamond\emptyset$ if and only if the input graph is $3$-colorable. This is claimed by the following result:

\begin{lemma} \label{th:succ3col-npalg}
Let $G=\{g_i=(a_i, b_i, c_i)\ | \ 1 \leq i \leq k\}$ be a boolean circuit encoding a graph, and let $\psi$ be the \npalg\ query built as described above.
The expression for \fail\ in $\psi$ evaluates to the empty relation for a given extension of $\relation{G}_1, \ldots \relation{G}_k, \relation{COL}_1, \relation{COL}_2, \relation{COL}_3$ if and only if $\relation{G}_1, \ldots \relation{G}_k$ correctly encode the circuit $G$ and $\relation{COL}_1, \relation{COL}_2, \relation{COL}_3$ represent a valid coloring of the input graph. Thus, $\fail\Diamond\emptyset$ if and only if the input graph is $3$-colorable.
\begin{proof} 
The boolean circuit $G$ is translated into $k$ guessed relations $\relation{G}_1, \ldots \relation{G}_k$. The correctness of the translation is claimed by Lemma~\ref{th:succcirc-NPAlg}. 
Moreover, the \guess\ part of query $\psi$ generates also all possible extensions for three more guessed relations, i.e., $\relation{COL}_1$, $\relation{COL}_2$, $\relation{COL}_3$. As discussed in Section~\ref{sec:coloring}, the expression for $\fail\relation{\_PARTITION} \cup \fail\relation{\_COLORING}$ evaluates to the empty relation if and only if $\relation{COL}_1$, $\relation{COL}_2$, $\relation{COL}_3$ define a valid coloring of the graph.
\end{proof}
\end{lemma}

\noindent
From previous lemmas, it follows the proof of Theorem~\ref{thm:combinedcompl} that states the combined complexity of \npalg:

\begin{proof}[Proof of Theorem~\ref{thm:combinedcompl}]
Immediate, from Lemma~\ref{th:succ3col-npalg} and from the NE-completeness of the Succint $3$-coloring problem \cite{kola-papa-91}.
\end{proof}

\section*{Acknowledgements}
This research has been supported by 
MIUR (Italian Ministry for Instruction, University, and Research) under the
FIRB project ASTRO (Automazione
dell'In\-ge\-gne\-ria del Software basata su Conoscenza), and
under the COFIN project ``Design and development of a software system for the
specification and efficient solution of combinatorial problems, based on
a high-level language, and techniques for intensional reasoning and local
search''. 
Special thanks are due to the anonymous reviewers, whose comments and suggestions lead to significant improvements of this paper.

\bibliographystyle{acmtrans}
{
\def\BIBPATH{../../../../bib/Toni}
\bibliography{%
%
%  BIB STRINGS
%
\BIBPATH/ToniLongFrequentStrings,%
\BIBPATH/ToniStringsForPublications,%
\BIBPATH/ToniLongEvents,%
%
%  BIB DATABASES
%
\BIBPATH/ToniUpdate,%
%
% BIB CONFERENCE EVENTS
%
\BIBPATH/ToniConferenceEventsEntries%
}

}

\end{document}